\documentclass{article}
\usepackage{booktabs}
\usepackage{amssymb}
\usepackage{booktabs}
\usepackage{pifont}
\usepackage{algorithm}
\usepackage{tabularx}
\usepackage{algpseudocode}
\usepackage{amsmath}
\usepackage{graphicx}
\usepackage{wrapfig}
\usepackage{caption}
\usepackage{subcaption}
\graphicspath{{figs/}}

\usepackage[preprint]{corl_2026} 
\hypersetup{pdftitle={PhySA: Decoupling Aesthetic and Physical Rewards for Humanoid Locomotion}, pdfauthor={Yunwan Fu, Sicheng He, Jiageng Mao, Yue Wang}}
\makeatletter
\renewcommand{\@noticestring}{Accepted to the CoRL 2026 Workshop on R2RL: Roadmap to Sample-Efficient Real-World Reinforcement Learning.}
\makeatother

\title{PhySA: Decoupling Aesthetic and Physical Rewards for Humanoid Locomotion}

\author{
  Yunwan Fu$^{1}$ \quad Sicheng He$^{2}$ \quad Jiageng Mao$^{2}$ \quad Yue Wang$^{2}$ \\
  $^{1}$Shanghai Jiao Tong University \quad $^{2}$University of Southern California \\
  \texttt{asterrose@sjtu.edu.cn}
}

\begin{document}
\raggedbottom
\maketitle
\vspace{-32pt}
\begin{figure}[H]
  \centering
  \includegraphics[width=0.97\linewidth]{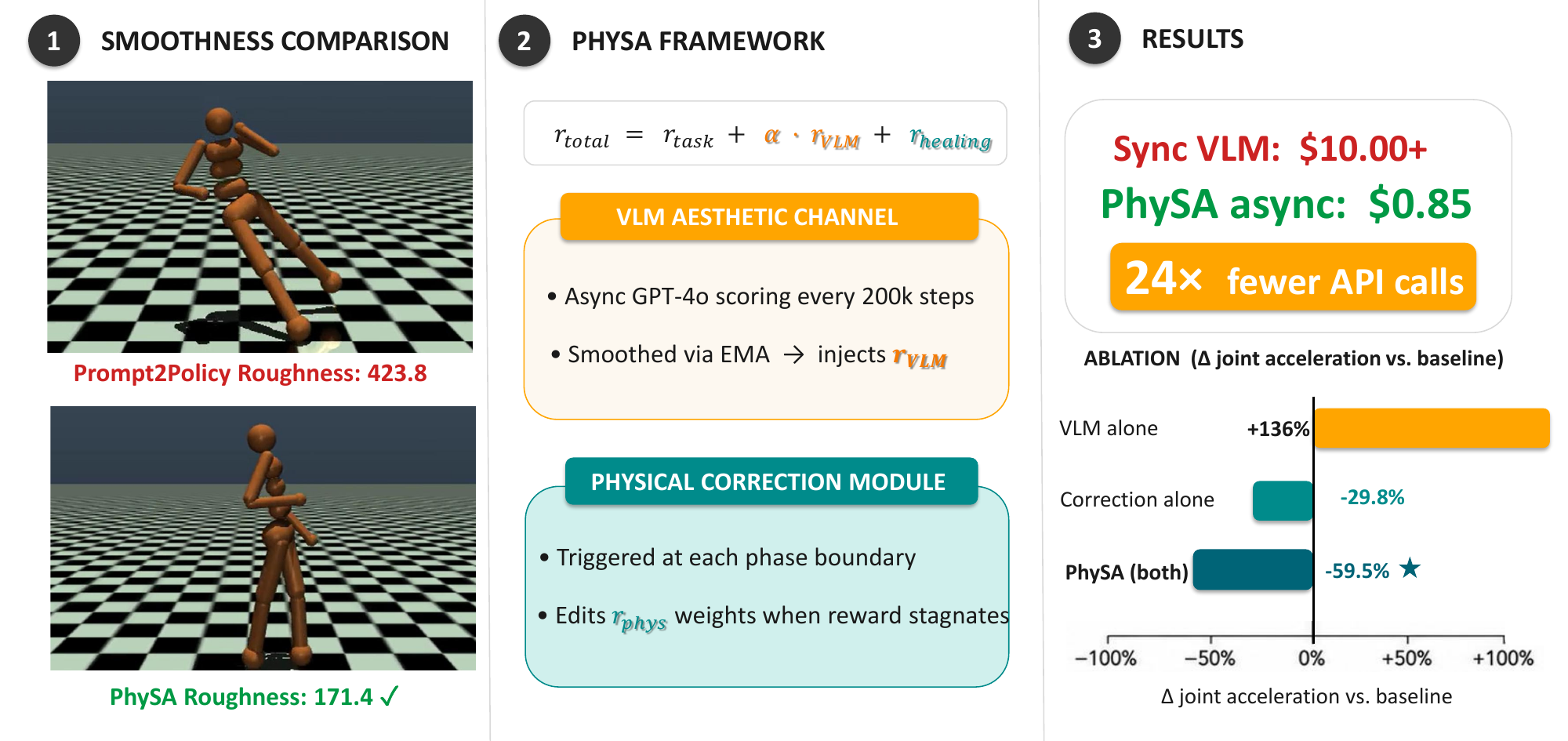}
  \caption{\textbf{Left:} In our main training run (a single run), PhySA reduces joint acceleration by $59.5\%$ relative to Prompt2Policy at the same forward-velocity target; a three-seed replication gives a smaller, non-significant $7.2\%$. \textbf{Center:} The two channels pull in opposite directions: the aesthetic channel alone \emph{raises} joint acceleration by $136\%$; the correction module alone reduces it by $29.8\%$. \textbf{Right:} PhySA issues $24{\times}$ fewer VLM queries than a synchronous per-update scorer would (estimated from the number of policy updates; measured PhySA cost \$$0.85$/10M steps); the smoothness gain does not carry over to other simulators, even after rescaling the reward weights.}
  \label{fig:overview}
\end{figure}


\begin{abstract}
Humanoid locomotion policies trained without an explicit smoothness objective tend to produce jerky, high-energy gaits that stress actuators. Vision-language models (VLMs) offer an appealing reward source: given a rendered rollout, a VLM returns an \emph{aesthetic quality} score on $[0,10]$ without task-specific feature engineering. We find, however, that aesthetic preference and physical smoothness point in opposite directions: across four production VLMs, clips with higher joint acceleration receive higher aesthetic scores (mean Pearson $r{=}{+}0.66$). We study \textbf{PhySA} (Physical Smoothness and Aesthetics), which keeps the two on separate, independently removable reward channels: a sparse, asynchronous VLM score and a phase-triggered correction channel that adjusts physical penalty weights from simulator state. Ablations attribute the smoothness gain to the correction channel. On MuJoCo Humanoid-v5 at 2.6M steps, PhySA lowers joint acceleration by $26$ to $32\%$ relative to vanilla PPO, Text2Reward, and Eureka (three seeds), at the cost of shorter episodes. In a single extended run it lowers joint acceleration by $59.5\%$ relative to Prompt2Policy, while a three-seed replication shows a smaller, non-significant reduction ($7.2\%$). The gain does not transfer to IsaacLab or the Unitree G1, even after rescaling the smoothness term.
\end{abstract}



\section{Introduction}
\label{sec:intro}

Smooth motion is a practical requirement for deploying humanoid robots on hardware. High joint velocity accelerates actuator wear, increases energy draw, and produces trajectories that low-level servo controllers struggle to track faithfully~\cite{radosavovic2024real,zhang2025achieving,fu2024humanplus,he2024omnih2o,muzio2022deep}. Yet locomotion policies trained without an explicit smoothness objective consistently learn jerky, high-energy gaits. Encoding smoothness alongside forward progress in a single reward requires careful weight balancing: a large smoothness coefficient tends to destabilize early training, while a small one leaves joint acceleration $\|\ddot{\mathbf{q}}\|_2$ uncontrolled.

Vision-language models (VLMs) trained on human motion video offer an appealing alternative reward source: given a rendered rollout, a VLM returns a continuous aesthetic quality score on $[0,10]$ without task-specific feature engineering~\cite{mahmoudieh2022zero,kwon2023reward,wang2024rl,chen2025era}. Prompt2Policy~\cite{prompt2policy2026} and related work~\cite{hu2023look} demonstrate this in practice by querying a VLM at every training update and incorporating the score directly into the per-step reward. LLM-synthesized reward functions~\cite{ma2024eureka,xie2024text2reward} take a different route: reward code is written once from a natural-language description, avoiding in-loop queries, but the resulting structure is fixed and cannot respond to failure modes that only emerge during training. Smoothness-specific regularizers~\cite{mysore2021regularizing,lee2024gradient} encode the objective directly but require per-robot weight tuning and can destabilize early training if applied too aggressively.

One might hope that a VLM aesthetic score could serve as a jerk proxy, sidestepping hand-crafted penalties altogether. The data say otherwise. Across four production VLMs, jerkier, more energetic motion consistently scored higher; these models appear to read large joint excursions as confident locomotion and smooth, low-amplitude motion as stiff or hesitant. Using the aesthetic score as a smoothness signal would therefore push the policy toward higher joint acceleration, not lower. VLMs can, in principle, judge gait qualities such as knee flexion timing and swing symmetry that per-step losses do not encode, so the aesthetic signal may still be worth keeping, but not as a proxy for smoothness. Keeping aesthetic scoring and physical smoothness on separate channels lets each be measured and adjusted without disturbing the other.

We propose PhySA (Physical Smoothness and Aesthetics), a reward engineering framework for humanoid locomotion built around two separate channels: an asynchronous VLM aesthetic scorer and a phase-triggered physical correction module. The channels share no reward terms, so each can be ablated or tuned without touching the other. In a single extended training run on MuJoCo Humanoid-v5~\cite{todorov2012mujoco,brockman2016openai}, PhySA reduces joint acceleration $\|\ddot{\mathbf{q}}\|_2$ by $59.5\%$ relative to Prompt2Policy~\cite{prompt2policy2026}, with $43\%$ lower mechanical power and $57\%$ longer episodes; a three-seed from-scratch replication shows a smaller, non-significant reduction ($7.2\%$, $p{=}0.45$). Against vanilla PPO, Text2Reward, and Eureka at 2.6M steps (three seeds), PhySA lowers joint acceleration by $26$ to $32\%$ while producing shorter episodes. The VLM channel in isolation raises joint acceleration by $136\%$, and ablations attribute the smoothness gain to the correction channel. The smoothness gain does not carry over to IsaacLab or the Unitree G1, even after rescaling the smoothness term; a timestep experiment within MuJoCo points to the numerical integration scheme~\cite{todorov2014convex,song2001analysis,featherstone2008rigid} as a major source of this gap.

Our main contributions are:
\begin{itemize}
  \item A finding that VLM aesthetic scores and joint-acceleration roughness are positively correlated across four production VLMs (mean Pearson $r{=}{+}0.66$), so a VLM aesthetic score should not be used as a proxy for physical smoothness.
  \item \textbf{PhySA}, a reward design in which an asynchronous VLM aesthetic scorer and a phase-triggered physical correction module operate on two separate reward channels, each removable without affecting the other; ablations show the smoothness gain comes from the correction channel, and a timestep experiment points to physics integration as the main obstacle to cross-simulator transfer.
\end{itemize}


\section{Related Work}
\label{sec:related work}

\paragraph{Vision-language rewards for policy learning.} LIV~\cite{ma2023liv} and RoboCLIP~\cite{sontakke2023roboclip} build shared embedding spaces between video and language, computing dense per-step shaping rewards by projecting rollout frames into a joint vision-language space; these methods target task completion and goal alignment but do not monitor physical smoothness. Prompt2Policy~\cite{prompt2policy2026} and related work~\cite{hu2023look} query GPT-4-class VLMs on rendered rollouts at every policy update, incorporating the returned score directly into the per-step reward. This synchronous design blocks the training loop at each update and incurs API costs proportional to training length (roughly 1,220 calls per 10M steps). PhySA instead dispatches scoring calls asynchronously at one call per 200k steps (50 per 10M steps, $24\times$ fewer), and separates aesthetic scoring from smoothness correction into distinct reward terms so that each can be ablated independently.

\paragraph{LLM-synthesized reward functions.} Eureka~\cite{ma2024eureka} uses iterative LLM prompting~\cite{wei2022chain} to synthesize reward code from natural-language task descriptions, then evaluates and refines candidates over multiple rounds; once the best candidate is selected, the reward structure is frozen for the duration of training and cannot be revised in response to unexpected policy behaviors that emerge. Text2Reward~\cite{xie2024text2reward} follows a similar single-shot synthesis approach with the same structural limitation. PhySA addresses this by monitoring training statistics at phase boundaries and making targeted edits to individual reward terms when stagnation is detected, adapting the reward structure to what the policy is actually doing.

\paragraph{Smoothness in humanoid reinforcement learning.} Action-rate penalties~\cite{mysore2021regularizing,lee2024gradient} add fixed regularization terms to discourage rapid action changes; gradient-surgery projections~\cite{yu2020gradient} project conflicting gradients into a feasible subspace at optimization time; multi-objective formulations~\cite{van2014multi,pirotta2015multi,zitzler2007hypervolume,zhang2023hypervolume} treat smoothness and task performance as separate objectives with a fixed Pareto weighting; imitation-based motion priors~\cite{pastor2009learning,araujo2025retargeting} impose smoothness implicitly by training the policy to match pre-recorded reference trajectories. All of these approaches commit to a fixed reward structure or motion prior before training begins and are unable to adjust their objectives once training is underway. Applied too early, a fixed jerk penalty tends to destabilize learning; applied too late, it fails to dislodge an already entrenched high-jerk gait. PhySA instead monitors training statistics at phase boundaries and edits individual reward weights in response to what the policy is actually doing (Sec.~\ref{sec:module_iso}).

\paragraph{VLM evaluation biases.} VLMs used as evaluators carry well-documented preference biases that can diverge from the intended target~\cite{zheng2023judging,hashemi2024llm,shi2025judging,ye2025justice}. In text-generation settings these biases (length preference, self-consistency, positional effects) are well characterized; in robotics they are rarely measured. We find that across four production VLMs, jerkier motion clips receive higher aesthetic scores (mean Pearson $r{=}{+}0.66$, ${\sim}100$ paired episodes per model; Spearman inter-model agreement $[+0.49, +0.94]$, mean $+0.76$). The VLMs appear to read energetic limb motion as confident locomotion, the same heuristic that makes human raters prefer action in sports video. Since this bias runs opposite to physical smoothness, using the VLM score as a jerk proxy produces the wrong gradient. The two objectives need separate channels, not a single conflated score.

\section{Method}
\label{sec:method}

\begin{figure}[t]
  \centering
  \includegraphics[width=\linewidth]{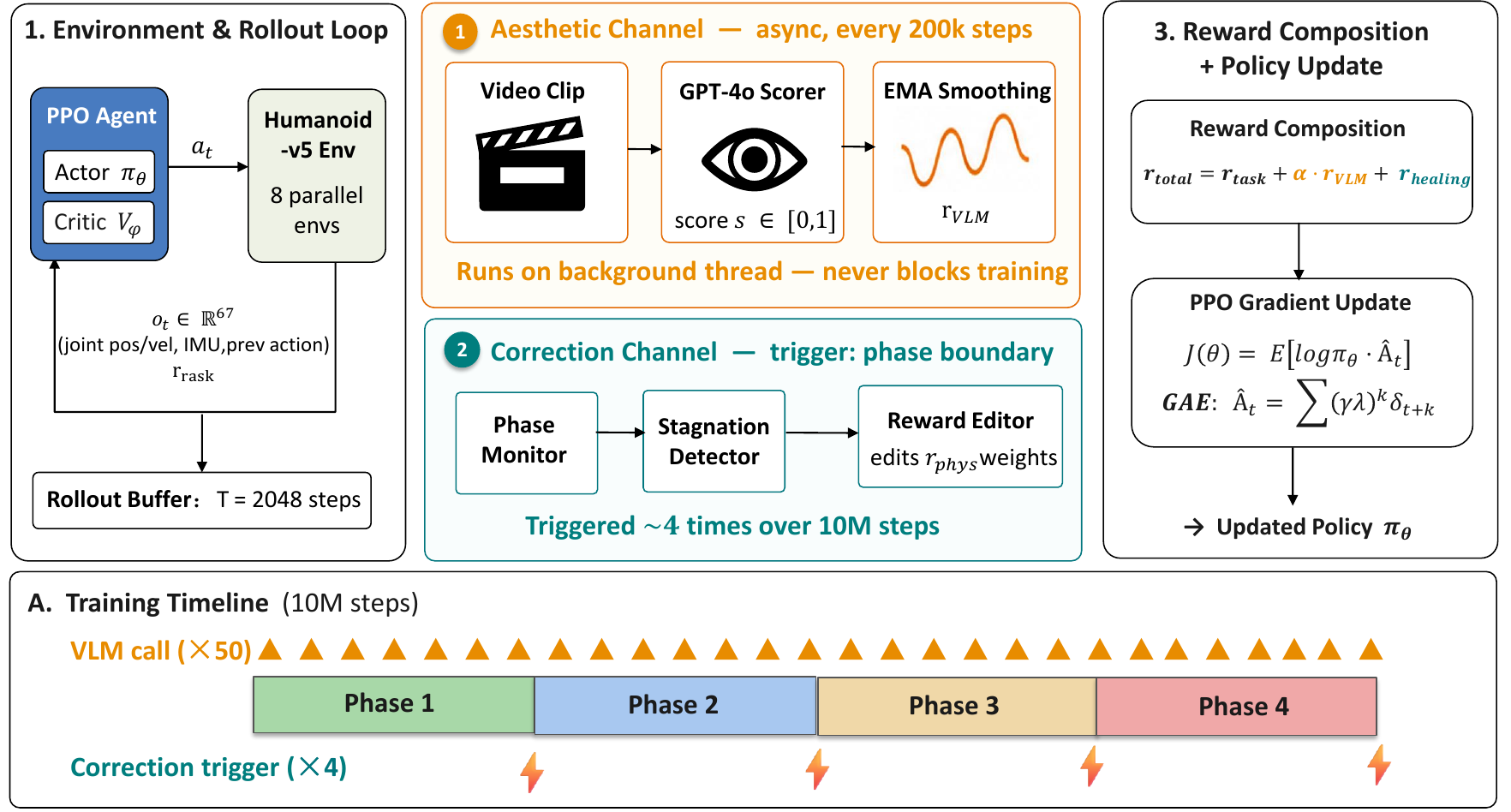}
  \caption{\textbf{PhySA end-to-end architecture.} \textit{Left:} The PPO agent interacts with MuJoCo Humanoid-v5; the total reward is Eq.~\ref{eq:phyrsa_total}. \textit{Center top (amber):} The aesthetic channel dispatches asynchronous GPT-4o scoring calls every $200\mathrm{k}$ steps on a background thread, smooths scores via EMA, and injects $r_{\text{VLM}}$ without blocking the rollout loop. \textit{Center bottom (teal):} The correction channel triggers at each phase boundary, checks for stagnating reward axes, and edits $r_{\text{phys}}$ weights accordingly. \textit{Right:} Reward composition and PPO gradient update. \textit{Bottom:} Training timeline showing 50 VLM calls (amber triangles) and 4 correction triggers (teal bolts) over $10\mathrm{M}$ steps.}
  \label{fig:pipeline}
\end{figure}

VLM-based reward has become a near-default choice in embodied AI, yet what a vision-language model evaluates from a rendered clip is appearance, not dynamics. A single frame contains no acceleration term; a VLM's notion of ``effort'' is therefore inherited from human video, where large limb excursions read as confident, purposeful motion. This learned prior is misaligned with physical smoothness, and the misalignment is not idiosyncratic to one model: all four production VLMs we evaluate rate jerkier rollouts more favorably (Sec.~\ref{sec:intro}). The implication for reward design is a clean separation of responsibility. Properties legible within a single frame (posture, limb coordination, bilateral symmetry) are well suited to VLM scoring. Properties defined by the trajectory's time-derivatives, such as joint acceleration, are not observable from a still frame and must instead be computed from simulator state; conflating the two into a single scalar would let the appearance-based signal dominate, since only that signal is actually supervised by the VLM.

A further complication is that the physical-side weight cannot be set a priori. Its effective scale is coupled to the simulator's integration timestep: within MuJoCo alone, varying $\Delta t$ shifts the same joint-acceleration statistic by close to $5\times$ (Sec.~\ref{sec:engine_gap}), and it also shifts with the policy's competence over the course of training, since a penalty that is negligible early on can come to dominate once the policy's control becomes more energetic, or the reverse. This is not a matter of insufficient tuning effort: a weight suitable at initialization has no guarantee of remaining suitable at convergence, so no offline sweep can be expected to identify one setting that holds throughout training. We instead leave this weight unspecified in advance and adjust it online at phase boundaries by monitoring training statistics (Sec.~\ref{sec:phys_channel}), while the VLM channel is held at a fixed weight throughout, responsible only for the appearance-level signal it can reliably supervise.

PhySA runs two reward channels on top of a standard Proximal Policy Optimization (PPO)~\cite{schulman2017proximal} setup implementing this division; Fig.~\ref{fig:pipeline} shows the complete system. The per-step reward is
\begin{equation}
  r_{\text{total}}(t) \;=\; r_{\text{task}}(t) \;+\; \alpha\,r_{\text{VLM}}(t) \;+\; r_{\text{phys}}(t) \;+\; r_{\text{GMP}}(t),
  \label{eq:phyrsa_total}
\end{equation}
where $r_{\text{task}}$ is a hand-crafted locomotion reward (forward velocity, alive bonus, height penalty, actuator cost), $\alpha\,r_{\text{VLM}}$ is the \emph{aesthetic} channel, $r_{\text{phys}}$ is the \emph{physical correction} channel, and $r_{\text{GMP}}$ is a Gaussian motion prior fit to motion-capture data~\cite{mahmood2019amass}, held constant across all conditions. $r_{\text{task}}$ covers the basic locomotion objective; $r_{\text{VLM}}$ is meant to reflect holistic gait qualities (knee use, swing symmetry, postural stability) judged from rendered clips (in its current form a global scalar, Sec.~\ref{sec:vlm_channel}); $r_{\text{phys}}$ enforces joint-acceleration smoothness, which VLM scoring actively opposes; and $r_{\text{GMP}}$ keeps the gait in a naturalistic range. Routing the aesthetic and correction signals through separate terms means either can be switched off cleanly (Sec.~\ref{sec:module_iso}).

\subsection{Task reward}
\label{sec:task_reward}

To isolate the effects of the two added channels, we carry over Prompt2Policy's~\cite{prompt2policy2026} task reward without modification: a Gaussian velocity-tracking term at $v^*{=}0.5\,$m/s, an alive bonus conditioned on torso height remaining within the standing band, a quadratic height penalty, and an L2 actuator cost. Any gait difference between PhySA and the baseline therefore originates in the two added channels, not in task-reward changes.

\subsection{Aesthetic reward channel}
\label{sec:vlm_channel}

Forward velocity and energy cost are straightforward to encode as per-step scalar penalties. Gait appearance qualities such as knee flexion timing, swing symmetry, and postural stability are harder to capture this way, as they depend on the trajectory as a whole rather than any single timestep. A VLM trained on large-scale human video can assess these holistically from a rendered clip, without task-specific feature engineering. The aesthetic channel (amber path in Fig.~\ref{fig:pipeline}) dispatches all scoring calls asynchronously on a background thread, so the training loop never blocks.

Every $200\mathrm{k}$ steps the trainer renders a $12$-frame clip and queries GPT-4o-2024-05-13~\cite{hurst2024gpt} (\texttt{temperature=0}). The prompt elicits a $[0,10]$ rating on three sub-axes (knee use, swing symmetry, center-of-mass tracking) plus an \texttt{overall} scalar; only \texttt{overall} enters the reward. Fifty calls over 10M steps is sufficient for the EMA to converge while keeping API spend under \$1. At ${\sim}2500$ steps/s on a single RTX~4090, each 200k-step interval spans roughly 80 s of wall time, comfortably longer than the 1--3 s API round-trip, so no rollout is ever delayed.

The aesthetic scalar is smoothed via an EMA and recentered,
\[
  \mathrm{vlm}_{\text{ema}}^{(t)} \;=\; 0.8\,\mathrm{vlm}_{\text{ema}}^{(t-1)} \;+\; 0.2\,s_t,
  \qquad r_{\text{VLM}} \;=\; \mathrm{vlm}_{\text{ema}}/10 - 0.5 \;\in\; [-0.5, +0.5],
\]
and enters Eq.~\eqref{eq:phyrsa_total} weighted by $\alpha$. The EMA keeps the per-step signal continuous between sparse queries and limits the impact of any single missed call. VLM scores agree across the four models we tested (Spearman correlation between model pairs: $[+0.49, +0.94]$, mean $+0.76$), though what the score measures semantically remains an open question. Appendix~\ref{app:alpha_sensitivity} confirms that smoothness improvements are driven by the correction channel, not the VLM signal. A policy trained specifically to maximize posture cues without a smoothness constraint does not outscore PhySA on this evaluator (Appendix~\ref{app:reward_hacking}), so the VLM is not simply rewarding jerk for its own sake. We fix $\alpha{=}0.5$; sensitivity across $\alpha\!\in\![0.2,0.8]$ is in Appendix~\ref{app:alpha_sensitivity}.

\paragraph{What this channel can and cannot do.} Because $r_{\text{VLM}}$ is updated once every 200k steps and added unchanged to every step until the next query, it does not depend on the current state or action. For example, a score of $6$ gives every step in the next interval the same $+0.05$ (with $\alpha{=}0.5$), whatever the robot's posture. Under the Humanoid-v5 termination rule, a constant per-step term changes only how much the policy values survival, much like an adjustment to the alive bonus; it cannot reward a specific posture or gait directly.

\subsection{Physical reward channel}
\label{sec:phys_channel}

The physical correction channel (teal path in Fig.~\ref{fig:pipeline}) applies phase-triggered edits to penalty terms in $r_{\text{phys}}$, targeting joint acceleration $\|\ddot{\mathbf{q}}\|_2$. Training is divided into four sequential phases of $2.5\mathrm{M}$ steps, each warm-started from the previous checkpoint. At each phase boundary, a criterion checks recent training logs across three axes (postural stability, energy efficiency, and episode reward) and fires a targeted reward edit when any axis regresses without a compensating gain elsewhere, or when all three stagnate simultaneously. A trigger raises actuator cost coefficients on high-velocity joints, removes a counterproductive penalty term, or leaves the reward unchanged; across both production runs, the criterion triggered at 3 of 8 phase boundaries ($37.5\%$). Phase-level triggering is preferred over step-level monitoring because short-window statistics primarily reflect gradient noise; the $2.5\mathrm{M}$-step interval gives the policy sufficient time to adapt before the next evaluation. Full implementation details are in Appendix~\ref{app:vlm_bridge}.

\section{Experiments}
\label{sec:experiments}

The experiments address three questions. \textbf{RQ1:} Does PhySA improve smoothness and task performance jointly, and what is the mechanism? \textbf{RQ2:} Which channel drives the smoothness gain, and can either be removed without loss? \textbf{RQ3:} Does the reward transfer across simulators and robot bodies, and if not, what breaks?

\subsection{Setup}
\label{sec:setup}

All conditions use Stable Baselines 3 (SB3)~\cite{raffin2021stable} PPO with identical hyperparameters (Appendix~\ref{app:training_setup}) on MuJoCo Humanoid-v5~\cite{akki2025benchmarking}; Prompt2Policy~\cite{prompt2policy2026} is the primary baseline. Each condition runs $N\!\in\!\{3,5,8,10,30\}$ seeds; cells with $N{<}8$ are directional only. Paired $t$-tests use false discovery rate (FDR) correction (Benjamini-Hochberg)~\cite{benjamini1995controlling,benjamini2001control} at $\alpha{=}0.05$; effect sizes are reported as standardized mean differences~\cite{lakens2013calculating}. Physical smoothness is measured as mean joint acceleration $\|\ddot{\mathbf{q}}\|_2$, finite-differenced from joint position. Comparisons across evaluation protocols use $\Delta\%$ only.

\subsection{MuJoCo Humanoid-v5: smoother motion across all metrics (RQ1)}
\label{sec:main_result}

PhySA reduces mean joint acceleration $\|\ddot{\mathbf{q}}\|_2$ by $59.5\%$ over Prompt2Policy in a single training chain (Table~\ref{tab:headline_results}). In this chain each method continues training from its own $10\mathrm{M}$ checkpoint for four $2.5\mathrm{M}$-step phases; a from-scratch replication over three seeds shows a smaller, non-significant gap ($-7.2\%$, $p{=}0.45$), so the headline magnitude should be read as a single-run result. Relative to Prompt2Policy, the smoothness gain does not come at the cost of episode length: episodes are $57\%$ longer, per-step mechanical power drops $43.1\%$, and per-episode work falls $13.1\%$ even after accounting for the longer episode duration. Against vanilla PPO, Text2Reward, and Eureka, however, PhySA trades shorter episodes for lower joint acceleration (Appendix~\ref{app:sota_matrix}). Episodes are also short in absolute terms ($123.5$ steps, about $1.9$~s, for PhySA), so these measurements cover early locomotion rather than long-horizon gait quality.

\begin{table}[h]
\centering
\caption{PhySA vs.\ Prompt2Policy at $10\mathrm{M}$ steps, MuJoCo Humanoid-v5. Joint accel.\ $= \|\ddot{\mathbf{q}}\|_2$; power $P(t) = \tau^\top \dot{q}$; VLM score is GPT-4o overall aesthetic on $[0,10]$. \textit{Note:} VLM scores are drawn from the $\alpha$-sensitivity sweep and are not statistically distinguishable at higher seed counts (Appendix~\ref{app:alpha_sensitivity}); joint accel., ep.\ length, power, and work are from a single 4-phase training chain in which each method continues from its own $10\mathrm{M}$ checkpoint.}
\label{tab:headline_results}
\setlength{\tabcolsep}{8pt}
\resizebox{\linewidth}{!}{%
\begin{tabular}{lccccc}
\toprule
Method & Joint accel.\ $\downarrow$ & VLM score $\uparrow$ & Ep.\ length $\uparrow$ & Power (W) $\downarrow$ & Work (J) $\downarrow$ \\
\midrule
Prompt2Policy         & $423.8$        & $4.81$         & $78.5$         & $999.6$        & $1195$ \\
\textbf{PhySA (ours)} & $\mathbf{171.4}$ & $5.74$ & $\mathbf{123.5}$ & $\mathbf{568.6}$ & $\mathbf{1038}$ \\
\bottomrule
\end{tabular}}
\end{table}

\begin{figure}[H]
\centering
\includegraphics[width=1\linewidth]{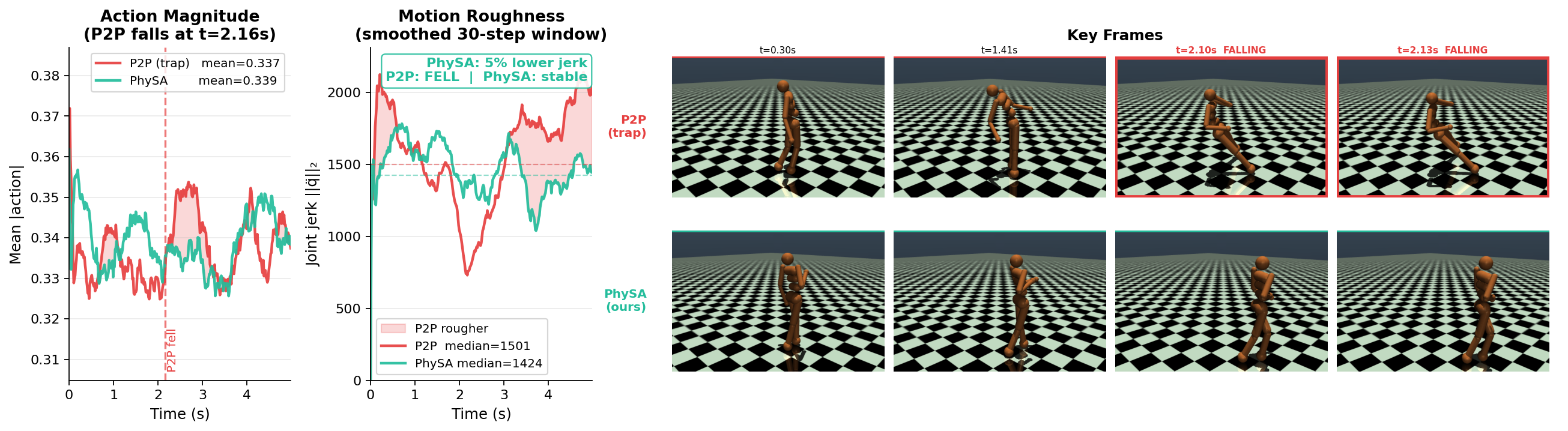}
\caption{\textbf{Trap reward (left) vs.\ PhySA corrected reward (right).} A reward augmented with $5.0\,\bar{|a|}+3.0\,\bar{|\Delta a|}$ drives the policy to instability: it falls at $t{=}2.16\,$s while the corrected policy completes the rollout. Action magnitude is equal across both ($\approx0.34$); instability stems from jerkier action transitions. Key frames at $t\in\{0.30,\,1.41,\,2.10,\,2.13\}$~s show torso height diverging to $z{=}1.06$~m (trap) vs.\ $z{=}1.25$~m (PhySA) at $t{=}2.10$--$2.13$~s.}
\label{fig:room_comparison}
\end{figure}

In a per-joint replay (Appendix~\ref{app:per_joint_energy}), the power saving is concentrated in two hip joints, which together account for $95.3\%$ of the reduction.
Torque norm rises $+6\%$ while joint-velocity norm falls $-35\%$: the policy achieves smoother motion through slower joints, not reduced actuator effort (Appendix~\ref{app:per_joint_energy}; Fig.~\ref{fig:room_comparison}).

\paragraph{Mechanism.} The correction module raises penalty coefficients on high-velocity joints; the policy responds by concentrating movement into a smaller set of joints and holding the rest nearly still. This specialization is quantified by per-joint action standard deviation, which drops from $0.353$ (module-free baseline) to $0.144$; per-joint action variation correlates with roughness across four ablation conditions at $r{=}{+}0.949$ (Appendix~\ref{app:five_way}). Smoothness is thus achieved through gait specialization rather than uniform deceleration, which explains why torque rises while velocity falls.

\subsection{Module isolation: identifying the source of the smoothness gain (RQ2)}
\label{sec:module_iso}

\begin{wrapfigure}{r}{0.46\linewidth}
\centering
\vspace{-0.5\baselineskip}
\includegraphics[width=\linewidth]{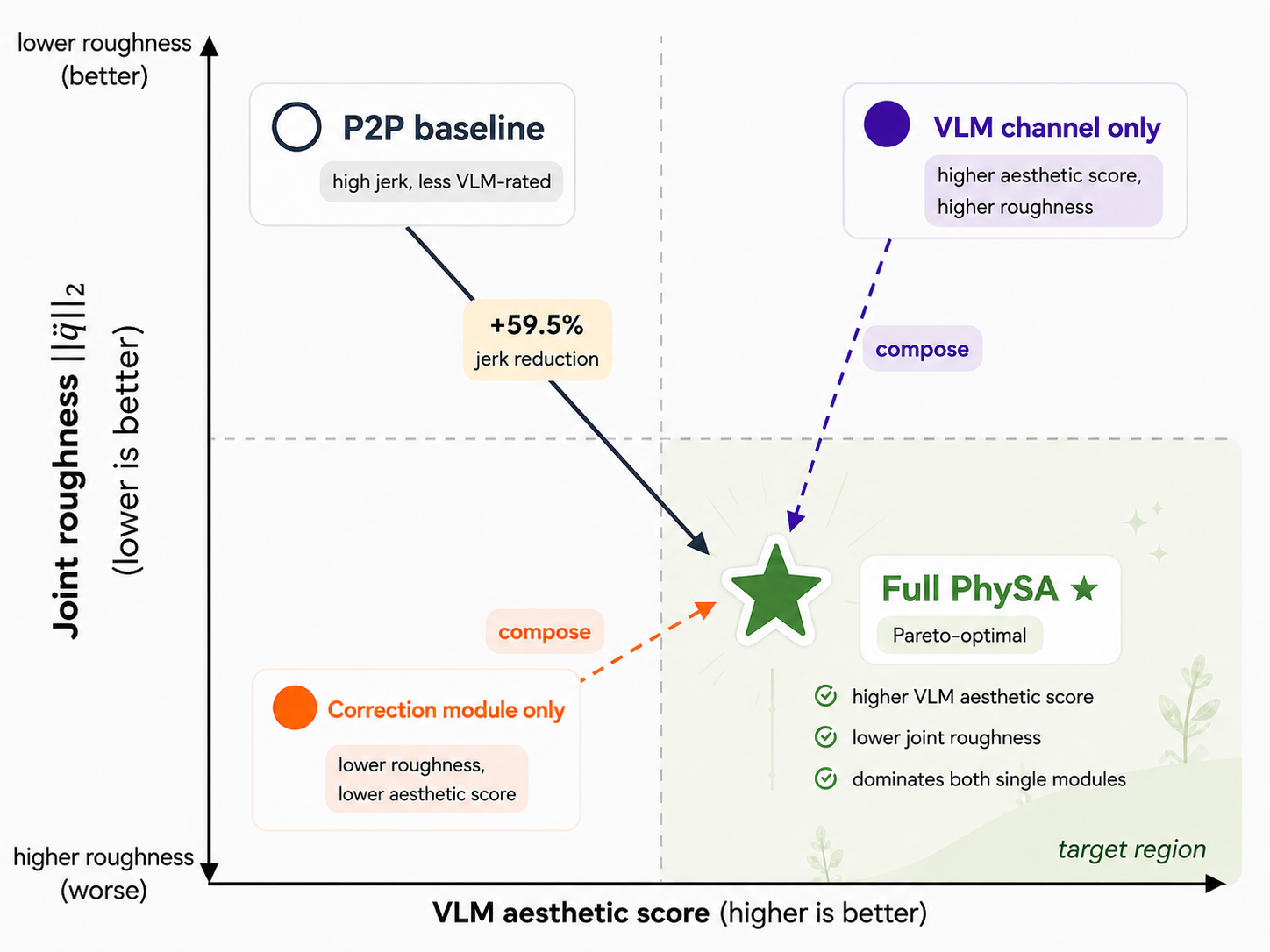}
\caption{Module isolation: VLM channel increases roughness; correction module reduces it.}
\label{fig:module_decoupling}
\vspace{-0.5\baselineskip}
\end{wrapfigure}
Running each channel in isolation confirms that the two pull in opposite directions on the smoothness axis. The VLM aesthetic channel alone increases roughness by $136\%$ while scoring higher on all four VLM evaluation dimensions ($N{=}8$; Appendix~\ref{app:vlm_robustness}). The correction module alone reduces roughness by $29.8\%$ but earns lower aesthetic scores ($N{=}50$). The combined PhySA result does not lie on the line connecting these two points (Fig.~\ref{fig:module_decoupling}): neither channel can substitute for the other. The positive correlation between roughness and aesthetic score is consistent across GPT-4o-2024-05-13, GPT-4o-2024-11-20, Claude Sonnet 4.6, and Gemini 2.5 Pro (Pearson $r$ range $+0.49$--$+0.82$; Appendix~\ref{app:vlm_robustness}); a preliminary check with LLaVA-NeXT-Interleave~\cite{li2024llavanextinterleave} showed the same directional result. In the combined stack, however, removing the VLM channel or varying $\alpha$ by $4{\times}$ has no measurable effect (Table~\ref{tab:five_way_main}; Appendix~\ref{app:alpha_sensitivity}), consistent with the state-independent form of $r_{\text{VLM}}$ (Sec.~\ref{sec:vlm_channel}).

\subsection{Diagnosing the transfer gap (RQ3)}
\label{sec:engine_gap}

Per-platform reward tuning is well-established in simulation-to-real transfer~\cite{peng2018sim,tobin2017domain}. The experiments here are not about demonstrating successful transfer; they are about identifying \emph{which component of the reward breaks and why}, a diagnosis that points at what must be fixed before hardware deployment. The short answer is that the per-step scale of the smoothness term shifts across simulators with the integration timestep, while the reward direction stays intact.

\begin{figure}[H]
\centering
\includegraphics[width=1\linewidth]{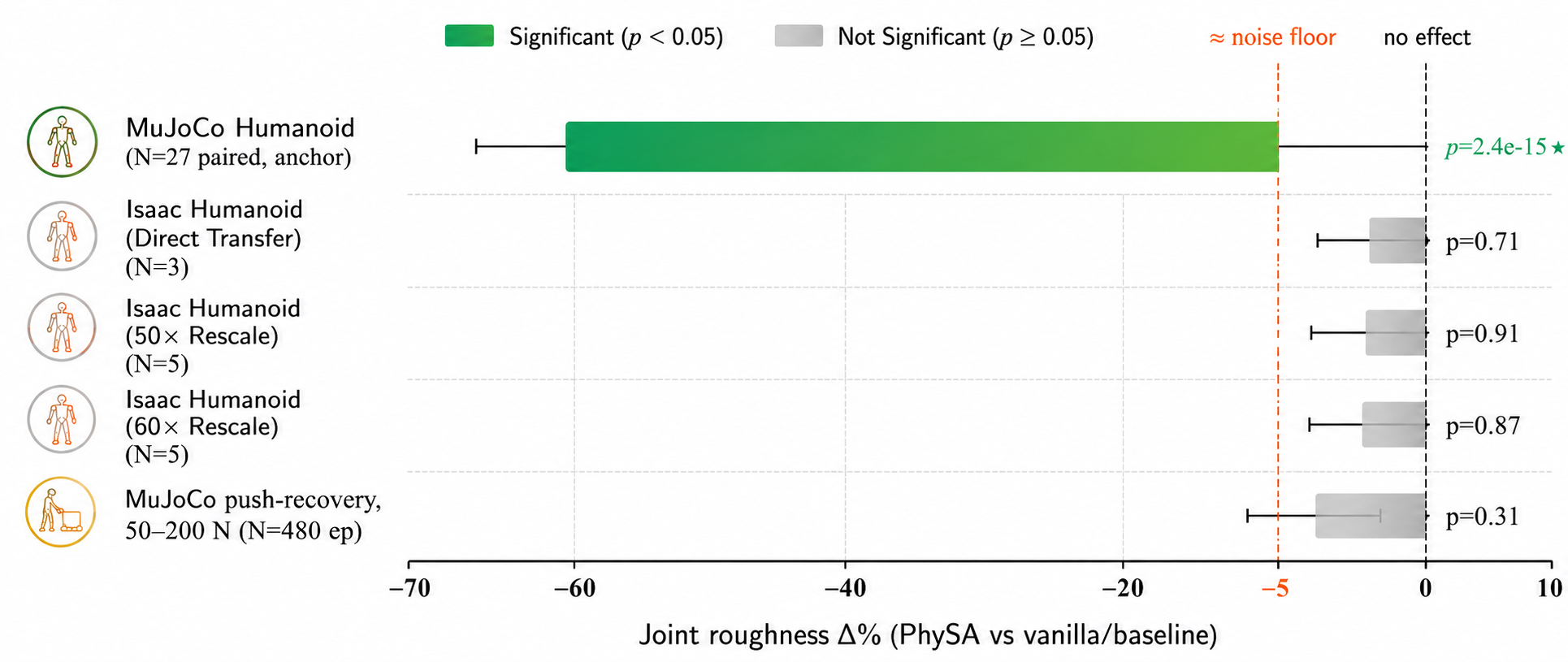}
\caption{\textbf{Smoothness gain does not transfer across simulators, even with rescaling.} Isaac: $-1.0\%$ ($p{=}0.91$, $N{=}5$, $50$--$60{\times}$ rescaling); G1: $-1.7\%$ (n.s., 38 settings). The MuJoCo anchor bar is a single training run, so its $p$-value does not reflect seed-to-seed variation (three-seed replication: $-7.2\%$, $p{=}0.45$). Conditions in Appendix~\ref{app:cross_engine}.}
\label{fig:cross_engine_bars}
\end{figure}

\paragraph{Isaac simulator (physics engine change only).} To isolate the effect of the physics engine while holding everything else constant, we transfer the PhySA policy to IsaacLab~\cite{makoviychuk2021isaac,mittal2023orbit} with the same humanoid task specification and the same robot body. Direct transfer gives a $-2.2\%$ smoothness gap, within noise. Empirically, reweighting the smoothness term by $50{\times}$--$60{\times}$ (chosen empirically to match the per-step magnitude of the smoothness term) keeps the gap at $-1.0\%$ to $-1.1\%$ ($p\!\in\!\{0.91,0.87\}$, $N{=}5$; Fig.~\ref{fig:cross_engine_bars}). This is a controlled experiment: only the physics engine changes. The Unitree G1 experiment below changes both simulator and embodiment simultaneously, so its null result cannot be attributed to either factor alone.

\paragraph{Isolating the integrator.}
To test whether the integration step is a major cause of the smoothness gap, we retrain in MuJoCo with $\Delta t$ forced to $1/60$~s ($N{=}3$) while holding all other settings fixed. Joint acceleration worsens by nearly $5{\times}$. At the standard MuJoCo timestep ($0.015$~s), each physics step internally integrates over five substeps, averaging out contact-spike noise; at the coarser step size those spikes are unsmoothed and dominate the $\|\ddot{\mathbf{q}}\|_2$ measure. The policy's control frequency (set by \texttt{frame\_skip}) is unchanged across both conditions: the policy receives the same number of action decisions per simulated second, and only the physics engine's internal integration resolution differs. This within-MuJoCo swap points to the integration scheme as a major source of the cross-simulator gap and confirms that the reward direction itself remains intact (Appendix~\ref{app:cross_engine}).

\paragraph{Unitree G1 robot.} Unlike the Isaac experiment, the G1 experiment changes both the simulator and the robot body simultaneously (a different physics engine and a different 23-joint embodiment), so any outcome cannot be attributed to either factor alone. One concrete miscalibration illustrates the scope of the problem: the Humanoid-v5 alive bonus fires while $z\!\geq\!1.1$~m, a threshold that never triggers on the G1 at its standing height of $0.793$~m, effectively inverting the survival signal. After fixing this threshold and running a 38-condition hyperparameter search, the smoothness improvement remains non-significant (mean $-1.7\%$, best $-4.8\%$; Appendix~\ref{app:g1_alive_bonus}). A G1 experiment inside MuJoCo, holding the simulator fixed and changing only the embodiment, would cleanly separate the two factors but was not run; this remains the most direct next step for attributing the G1 null result.


\section{Conclusion}
\label{sec:conclusion}

\textbf{PhySA} separates VLM aesthetic scoring from physical smoothness correction into two independent reward channels. The two objectives conflict: VLMs score jerkier motion higher, the VLM channel alone raises joint acceleration by $136\%$, and the correction module alone reduces it by $29.8\%$; in the combined stack the gain comes from the correction channel. A VLM aesthetic score should therefore not serve as a smoothness proxy; smoothness needs its own channel.

\textbf{Limitations and future work.} Our largest effect ($59.5\%$) comes from a single training run; a three-seed replication shows a non-significant $7.2\%$. Episodes are short (Sec.~\ref{sec:main_result}), and the VLM channel acts only through the value of survival (Sec.~\ref{sec:vlm_channel}), so a clip-level or state-conditioned VLM reward is a natural next step. The gain does not carry over to other simulators even after rescaling, and on the G1 a height threshold copied from Humanoid-v5 kept the alive bonus from firing until fixed (Appendix~\ref{app:g1_alive_bonus}); a G1 run inside MuJoCo is the natural next step.


\newpage
{\bibliography{example}}  


\clearpage
\appendix
\normalfont\rmfamily\mdseries\upshape\normalsize

\section*{Appendix}

\noindent
Reproducibility details are in Appendix~\ref{app:impl}; extended quantitative evidence in Appendix~\ref{app:extended_results}; scope boundaries in Appendix~\ref{app:scope}; and the cross-VLM aesthetic audit in Appendix~\ref{app:vlm_robustness}.

\begin{table}[H]
\centering
\caption{VLM-based reward methods. In-loop: sync = blocks training, async = background thread. Cost: API spend per $10\mathrm{M}$ steps. Smoothness: whether the method monitors a physical smoothness signal.}
\label{tab:related_work_matrix}
\setlength{\tabcolsep}{8pt}
\begin{tabular}{lllcc}
\toprule
Method & VLM role & In-loop & Cost (\$ / $10\mathrm{M}$) & Smoothness \\
\midrule
SayCan~\cite{ahn2022can}                              & planner         & no       & N/A          & no  \\
LIV / RoboCLIP~\cite{ma2023liv,sontakke2023roboclip}  & dense critic    & sync     & $> 10$       & no  \\
Prompt2Policy~\cite{prompt2policy2026}                & dense critic    & sync     & $> 10$       & no  \\
Eureka~\cite{ma2024eureka}                            & offline writer  & one-shot & $25$--$30$   & no  \\
Text2Reward~\cite{xie2024text2reward}                 & offline writer  & one-shot & $< 0.50$     & no  \\
\textbf{PhySA (ours)}                                 & scorer + corrector & async    & \textbf{0.85} & \textbf{yes} \\
\bottomrule
\end{tabular}
\end{table}

\section{Implementation Details and Reproducibility}
\label{app:impl}

\subsection{Aesthetic reward channel, physical reward channel, and training stabilization}
\label{app:vlm_bridge}

Every $200\mathrm{k}$ env steps the trainer renders a $12$-frame single-view MP4 (MPEG-4 video, $0.4$~s at $30$~fps, written via \texttt{imageio}) and dispatches a GPT-4o-2024-05-13 scoring call~\cite{hurst2024gpt} (\texttt{temperature=0}, \texttt{seed=42}) on a background \texttt{ThreadPoolExecutor}; the main PPO loop never blocks (measured wall-clock overhead $-4.1\%$ at $1\mathrm{M}$, Appendix~\ref{app:cost_audit}). The VLM returns a JSON (JavaScript Object Notation) response with three per-dimension aesthetic scores (knee redundancy, swing symmetry, COM tracking) and an \texttt{overall} scalar in $[0,10]$; only the overall feeds the reward channel via an EMA, $\mathrm{vlm}_{\text{ema}}^{(t)} = 0.8\,\mathrm{vlm}_{\text{ema}}^{(t-1)} + 0.2\,s_t$, mapped to $r_{\text{VLM}} = \mathrm{vlm}_{\text{ema}}/10 - 0.5 \in [-0.5, +0.5]$.

\paragraph{Phase-triggered modification.} At each $2.5\mathrm{M}$-step phase boundary the selection criterion (multi-criteria check: has any training axis regressed without a compensating gain on another?) reads the last $30$ PPO log rows and computes $\Delta\mathrm{stability}, \Delta\mathrm{energy}, \Delta\mathrm{ep\_mean}$. It triggers a targeted reward edit under either a stability regression ($\Delta\mathrm{stability}<-0.04$) or full stagnation (no metric improves by $\geq +0.01$); otherwise it proceeds to the next phase without modification. Across the two production runs described in Sec.~\ref{sec:experiments}, the criterion triggered at $3$ of $8$ phase boundaries ($37.5\%$).

\subsubsection{Axis-to-reward-term mapping}
\label{app:llm_judge_prompt}

The correction module identifies the underperforming control axis from recent training metrics via the multi-criteria stagnation detection procedure described in Appendix~\ref{app:vlm_bridge}, returning a label $\in \{\texttt{knee}, \texttt{swing}, \texttt{com}, \texttt{none}\}$. Depending on the identified axis, targeted modifications are applied to specific reward terms: a \texttt{knee} verdict raises the L2 actuator cost at the two knee indices (\texttt{ctrl\_cost}, indices $6$ and $10$), while \texttt{com} instead zeros the upright-balance penalty (\texttt{upright}). A \texttt{swing} verdict removes the asymmetric swing penalty (\texttt{swing\_trap}) rather than adjusting joint costs. If the criterion returns \texttt{none}, no term is modified and the currently active reward terms remain frozen for the subsequent phase.

The per-phase diagnostic logs record what the criterion computed at each boundary (selection criterion output, stagnation flag, axis decision). In the PhySA trace, the targeted reward term shifts from a knee-velocity group at phase 0 to a single hip-velocity axis by phases 2--3; the set of active modifications narrows over training as the reward edits take effect.

\paragraph{Stabilization patches.} Three additional components further stabilize the two-channel reward: EMA smoothing on $r_{\text{VLM}}$ ($\alpha{=}0.2$, variance ratio $0.816$ at $10\mathrm{M}$); an early-training guard that scales $\alpha_{\text{eff}}{=}0.2\alpha$ when the mean episode length falls below $100$ steps (fires $0/400$ times at $10\mathrm{M}$, $3/10$ times at $1\mathrm{M}$); and axis-restricted reward edits within $r_{\text{phys}}$ indexed by the selection criterion's identified weak axis $\in \{\texttt{knee, swing, com, none}\}$. Each component is active in a different regime, so removing any single one leaves the others functional.

\paragraph{Design constraints behind the cadence and EMA.} The scoring cadence of one call per $200\mathrm{k}$ environment steps reflects two constraints: at our throughput ($\sim\!2500$ FPS (environment steps per second) on a single RTX~4090), $200\mathrm{k}$ steps span roughly $80$~s of wall-clock, long enough to amortize the $1$--$3$~s API round-trip without blocking the rollout queue; and at the $10\mathrm{M}$-step horizon this gives $50$ score updates per run, enough for the EMA to settle and few enough to keep API spend below \$$1$. The EMA constant $\alpha{=}0.2$ on $r_{\text{VLM}}$ keeps inter-call swings to $\pm 0.2$ on a stable training run, so an isolated rate-limit fallback (counted as a missed score, not a zero) does not drive a visible perturbation; smaller values over-smooth the early-training signal, where the policy is still trading falling-jerk for steady-state jerk and the score range is wide. The $\alpha_{\text{eff}}{=}0.2\alpha$ early-training guard was added after an $\alpha{=}0.8$ pilot at $1\mathrm{M}$ steps in which the VLM, asked to grade a humanoid that was still falling, returned near-zero scores that the policy chased into a sit.

\subsection{Auxiliary modules: SAGE and GMP}
\label{app:auxiliary}

SAGE (Structured Agent for Goal Elicitation) is a three-stage LLM agent (intent elicitor $\to$ reward author $\to$ judge) that generates the initial $r_{\text{task}}$ template later adjusted by the correction module; its outputs are evaluated separately as single-shot benchmarks and do not affect the headline numbers. GMP (Gaussian Motion Prior) is a per-step reward fit to eight motion-capture categories~\cite{mahmood2019amass} (\texttt{walk\_slow}, $\ldots$, \texttt{stairs}, $K{=}2$--$6$ components, \texttt{covariance\_type="full"}; \texttt{sklearn.mixture.GaussianMixture}~\cite{pedregosa2011scikit}); it appears as one cell of the 5-way ablation (Section~\ref{app:five_way}).

\subsection{Training setup}
\label{app:training_setup}

\begin{table}[H]
\centering
\setlength{\tabcolsep}{10pt}
\caption{Stable Baselines~3 PPO hyperparameters, identical across every cell in the paper.}
\label{tab:ppo_hyperparams}
\begin{tabular}{ll}
\toprule
Hyperparameter & Value \\
\midrule
Optimizer                 & Adam \\
Learning rate             & $3 \times 10^{-4}$ \\
Discount $\gamma$         & $0.99$ \\
GAE (Generalized Advantage Estimation) $\lambda$             & $0.95$ \\
Parallel environments     & $8$ \\
Rollout length            & $1024$ \\
Epochs per update         & $4$ \\
Mini-batches per update   & $4$ \\
Clip coefficient $\epsilon$ & $0.20$ \\
Entropy coefficient       & $0.01$ \\
Gradient clipping         & $0.5$ \\
Policy network            & Multilayer perceptron (MLP) $[256, 256]$ \\
\bottomrule
\end{tabular}
\end{table}

All training uses SB3~\cite{raffin2021stable} PPO~\cite{schulman2017proximal} (Adam~\cite{kingma2014adam}, GAE~\cite{schulman2015high}, PyTorch~\cite{paszke2019pytorch}) with the configuration in Table~\ref{tab:ppo_hyperparams}, frozen across every cell in the paper. Compute: a single dedicated RTX~4090 with $n_{\text{envs}}{=}8$ MuJoCo workers; peak VRAM (video random access memory) $\sim\!1.1$~GiB; PhySA $10\mathrm{M}$ wall-clock $3.45\!\pm\!0.57$~h. All experiments use \texttt{seed=42} with deterministic offsets for cross-seed runs. Trained checkpoints cover three experimental tracks; the runtime environment is pinned to \texttt{stable\_baselines3 2.7.1}, \texttt{torch 2.10.0}, \texttt{gymnasium 1.2.3}, Python $3.13.9$.

\section{Extended Results}
\label{app:extended_results}

\subsection{Per-joint energy breakdown}
\label{app:per_joint_energy}

\begin{wrapfigure}[17]{r}{0.45\linewidth}
\centering
\vspace{-4.5\baselineskip}
\includegraphics[width=\linewidth]{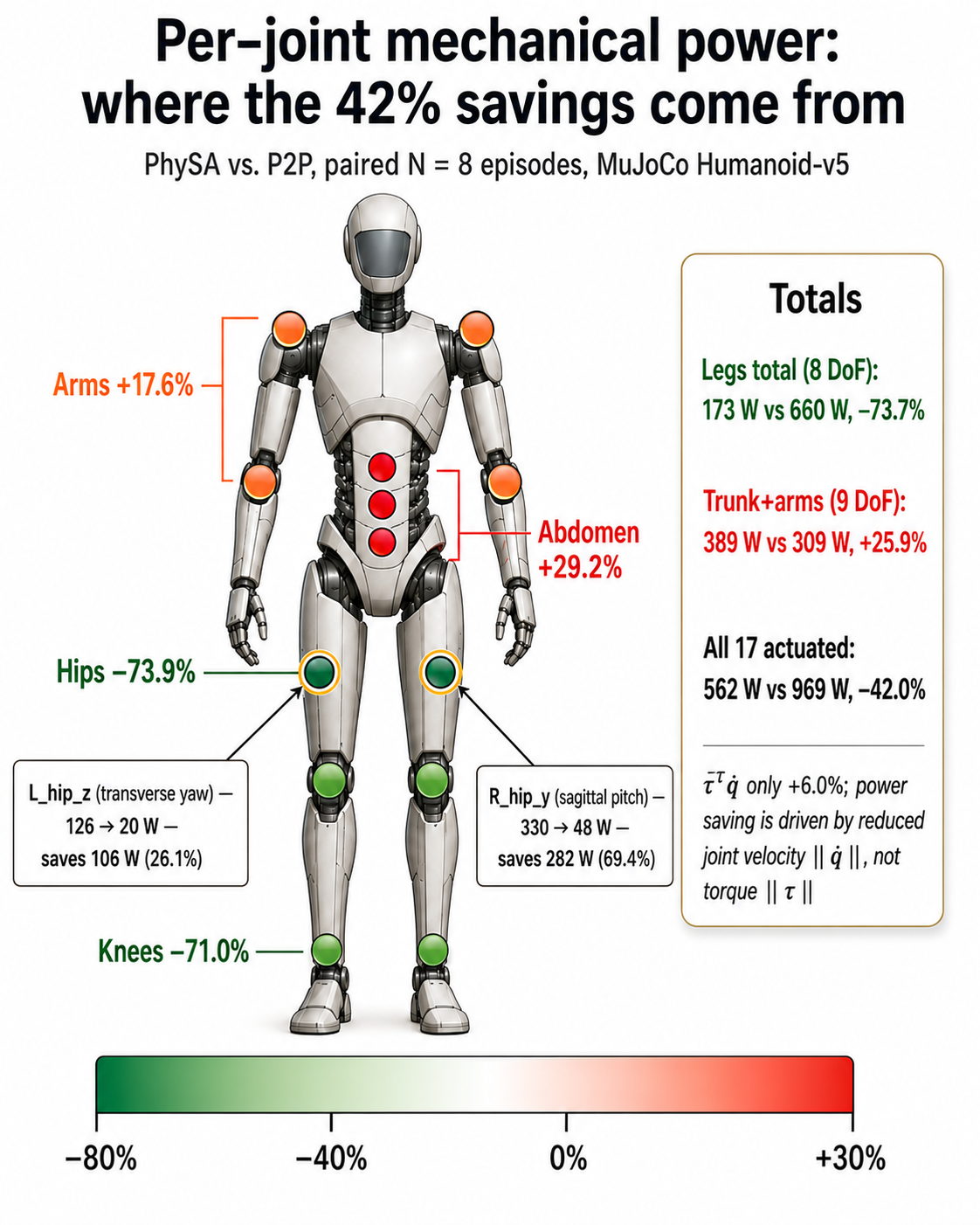}
\caption{\textbf{Per-joint power.} Legs $-73.7\%$; trunk/arms $+25.9\%$.}
\label{fig:per_joint_power}
\vspace{-0.5\baselineskip}
\end{wrapfigure}
\normalsize

Sec.~\ref{sec:main_result} reports $-43.1\%$ per-step mechanical power. A replay of the correction-module checkpoint bundle ($N{=}8$ paired episodes, seeds 42--45) decomposes this per joint group via $P_j(t){=}\tau_j(t)\dot{q}_j(t)$. The saving is not body-wide: legs drop $-486$~W while trunk and arms rise $+80$~W, with $95.3\%$ of the $406.6$~W net reduction carried by two hip joints, \texttt{R\_hip\_y} (sagittal stride) and \texttt{L\_hip\_z} (transverse yaw; Fig.~\ref{fig:per_joint_power}).

The $-43\%$ saving comes from reduced joint velocity, not torque: the per-step torque norm changes by only $+6.0\%$. Since actuator wear is dominated by joint velocity rather than peak torque, the $95.3\%$ hip share localizes the reduction. The trunk and arms use $+25.9\%$ more power, suggesting the upper body absorbs more of the stride's angular-momentum bookkeeping as the legs slow.

\subsection{Five-way single-knockout ablation}
\label{app:five_way}

We perform single-module knockouts from the full PhySA stack. The full-stack and no-correction-module conditions use $N{=}10$ paired seeds ($42$--$51$); the remaining four cells use $N{=}3$. The $N{=}10$ paired-$t$ on full-stack vs.\ no-correction-module ($\Delta\,\mathrm{ep\_len} = -9.6\%$, $p = 0.032$, effect size $d_z = -0.80$) is the only knockout comparison at $N{=}10$; the other cells are directional checks.
The complete five-cell knockout results are given below.

\begin{table}[H]
\centering
\normalsize
\caption{Five-way knockout. Full and w/o correction module use $N{=}10$ paired seeds; other cells are $N{=}3$ checks. VLM score is the GPT-4o overall aesthetic rating on $[0,10]$.}
\label{tab:five_way_main}
\begin{tabular}{@{}lccccc@{}}
\toprule
Cell & $N$ & Task return & VLM score & Ep.\ length \\
\midrule
\textbf{full}                   & $\mathbf{10}$ & $298.42 \pm 38.13$ & $5.87 \pm 1.07$ & $78.64 \pm 9.18$ \\
\texttt{w/o VLM}               & $3$           & $287.57 \pm 10.16$ & $6.43 \pm 0.12$ & $74.06 \pm 5.54$ \\
\textbf{\texttt{w/o correction}} & $\mathbf{10}$ & $\mathbf{279.70 \pm 22.41}$ & $\mathbf{5.19 \pm 1.89}$ & $\mathbf{71.09 \pm 5.49}$ \\
\texttt{w/o SAGE}              & $3$           & $370.92 \pm 32.35$ & $6.50 \pm 0.00$ & $73.76 \pm 5.27$ \\
\texttt{w/o jerk}              & $3$           & $315.86 \pm 34.25$ & $6.50 \pm 0.00$ & $78.45 \pm 7.16$ \\
\texttt{w/o GMP}               & $3$           & $311.17 \pm 46.26$ & $5.83 \pm 1.15$ & $79.83 \pm 7.12$ \\
\bottomrule
\end{tabular}
\end{table}
\normalsize

A per-joint bandwidth audit across four ablation conditions shows how the correction module reduces roughness: mean per-joint action standard deviation is $0.144$ vs.\ $0.353$ for a vanilla baseline ($41\%$ of vanilla), the minimum per-joint standard deviation drops to $3.8\times 10^{-6}$, the saturation rate reaches $92.8\%$, and mean $|a|$ is $0.384$. The policy concentrates movement in a small number of joints while holding the rest nearly fixed. Across the four conditions, per-joint action standard deviation and roughness correlate at $r{=}+0.949$.

\subsection{\texorpdfstring{$\alpha$}{alpha}-sensitivity: robust at $N{=}30$ paired}
\label{app:alpha_sensitivity}

A four-stage sequential protocol ($N{=}1\!\to\!3\!\to\!5\!\to\!30$) tests whether $\alpha$ is a robust hyperparameter. The $N{=}30$ paired-$t$ on $\alpha{=}0.2$ vs $\alpha{=}0.8$ across shared seeds $42$--$71$ is in Table~\ref{tab:snip_n30}: all five metrics fail to reject the null at $p<0.05$ with uniformly small effect sizes ($|d_z|<0.20$). Beyond the mean, $\alpha{=}0.8$ produces a measurable tail-narrowing on all five metrics (roughness interquartile range (IQR) tightens by $37\%$); the smoothness floor is set by the correction module and does not drift with the aesthetic weight (roughness change is $-2.0$ across the $4{\times}$ $\alpha$ range, $p{=}0.863$).

\begin{table}[H]
\centering
\normalsize
\caption{$\alpha$ sweep, $N{=}30$ paired. $10\mathrm{M}$-median protocol applied at the $3\mathrm{M}$ intermediate scale; compare only within this table.}
\label{tab:snip_n30}
\begin{tabular}{@{}lccccc@{}}
\toprule
Metric & $\alpha{=}0.2$ & $\alpha{=}0.8$ & $\Delta$ & $p$ & $d_z$ \\
\midrule
Task return                          & $593.5 \pm 52.9$  & $605.0 \pm 41.7$  & $+11.5$    & $0.362$ & $+0.169$ \\
Episode length                       & $119.1 \pm 10.7$  & $121.4 \pm 8.4$   & $+2.4$     & $0.353$ & $+0.172$ \\
Jerk p95                             & $109\,383$ & $109\,048$ & $-334$      & $0.736$ & $-0.062$ \\
Roughness $\|\ddot{\mathbf{q}}\|_2$   & $1139.7 \pm 46.6$ & $1137.7 \pm 49.7$ & $-2.0$     & $0.863$ & $-0.032$ \\
VLM overall                          & $4.83 \pm 0.76$   & $4.73 \pm 0.62$   & $-0.10$    & $0.537$ & $-0.114$ \\
\bottomrule
\end{tabular}
\end{table}
\normalsize

\subsection{Multi-method comparison and Pareto position}
\label{app:sota_matrix}

PhySA is compared with Eureka~\cite{ma2024eureka}, Text2Reward (T2R)~\cite{xie2024text2reward}, and a vanilla PPO baseline (PPO, task reward only) under a step-aligned protocol at $2.6\mathrm{M}$ training steps ($N{=}3$ seeds ${\times}$ 50 episodes, shared Humanoid-v5 / SB3 PPO infrastructure). All six smoothness comparisons against PhySA pass Benjamini--Hochberg correction at $\alpha{=}0.05$: $-25.7\%$ vs PPO ($t{=}{-}6.16$), $-26.6\%$ vs T2R ($t{=}{-}6.39$), $-31.7\%$ vs Eureka ($t{=}{-}7.64$) on mean smoothness score, with matching improvements at the 95th percentile. PhySA shows lower episode length and episode return against all three baselines (range $-37\%$ to $-63\%$), all statistically significant after correction. Eureka's 10M checkpoint shows a reward-exploitation failure (terminal displacement $|x|{<}0.003$~m; the synthesized reward does not penalize standing in place, so the policy oscillates and collects the alive bonus indefinitely, Appendix~\ref{app:eureka_collapse}), and T2R is dominated by PPO on both smoothness and episode length. Reward synthesis cost for Eureka/T2R is a one-time expense (Appendix~\ref{app:cost_audit}); PhySA's ongoing VLM cost is \$0.85 per 10M steps on matched compute. Per-method standard deviations across $N{=}3$ seeds: $\sigma_{\text{ep\_len}}{\le}27$, $\sigma_{\text{jerk}}{\le}51$.

\begin{figure}[H]
\centering
\begin{minipage}[t]{0.56\linewidth}
\vspace{0pt}
\centering
\small
\setlength{\tabcolsep}{3pt}
\captionof{table}{Multi-method comparison at $2.6\mathrm{M}$ steps (MuJoCo Humanoid-v5; $N{=}3\times50$ episodes). $^\ast$: PhySA smoothness wins. $^\dagger$: task-axis losses. Eureka did not produce forward walking (Appendix~\ref{app:eureka_collapse}); included for completeness.}
\label{tab:sota_matrix}
\begin{tabular}{@{}lrrrr@{}}
\toprule
Method & Joint accel. & Jerk p95 & Ep.\ len & Ep.\ ret \\
\midrule
Eureka                & 1015 & 1963 & 186.2 & 879.2 \\
Text2Reward           & \phantom{0}944  & 1810 & 116.4 & 625.9 \\
PPO                   & \phantom{0}933  & 1792 & 126.9 & 667.2 \\
\textbf{PhySA (ours)} & \textbf{\phantom{0}693$^\ast$} & \textbf{1423$^\ast$} & \phantom{0}73.7$^\dagger$ & 322.4$^\dagger$ \\
\bottomrule
\end{tabular}
\end{minipage}\hfill
\begin{minipage}[t]{0.41\linewidth}
\vspace{0pt}
\centering
\includegraphics[width=\linewidth]{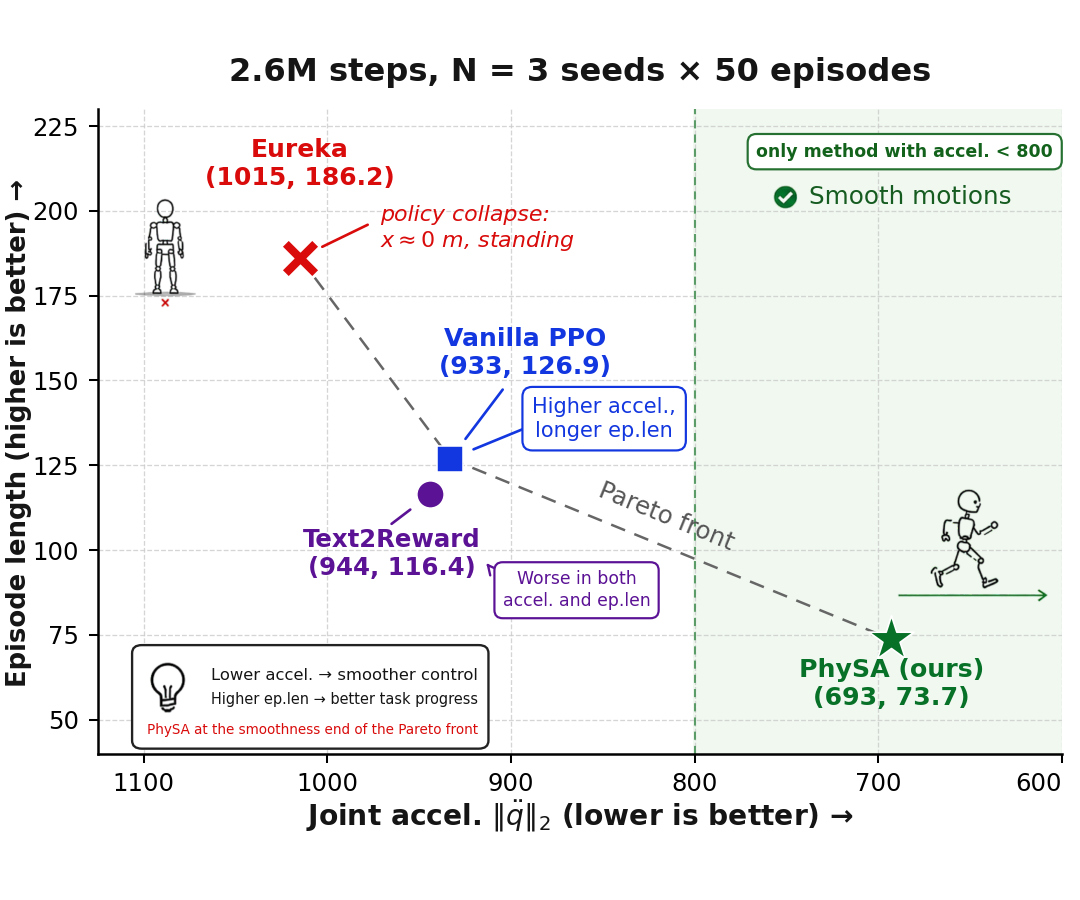}
\captionof{figure}{\textbf{Smoothness--episode-length Pareto front (Humanoid-v5, $2.6\mathrm{M}$).} PhySA holds the smoothness end.}
\label{fig:sota_pareto}
\end{minipage}
\end{figure}

\paragraph{Eureka anatomy.} Eureka's $10\mathrm{M}$ checkpoints oscillate in place while accumulating the alive bonus (legs $\overline{\|\dot q\|}{\approx}19$, terminal $|x|{<}0.003$~m, $z{\approx}1.40$~m; 3 seeds, 150 episodes), a reward-exploitation pattern~\cite{skalse2022defining,meyes2019ablation} that an offline reward synthesizer cannot detect during training, whereas the phase-triggered stagnation criterion catches it within one phase. PhySA's $2.6\mathrm{M}$ task-axis trade vs T2R/PPO corresponds to a trade-off index $\eta{\in}[0.61, 0.73]$; the matched $10\mathrm{M}$ correction-module vs.\ P2P comparison does not show this trade (Tab.~\ref{tab:headline_results}).

T2R is the only off-front cell on the (jerk, ep\_len) plane. PPO strictly dominates it on both axes: jerk $944{>}933$ and ep\_len $116{<}127$. PhySA is the only point at the smoothness end (Fig.~\ref{fig:sota_pareto}). The trade-off index $\eta = |\Delta_{\text{jerk}}\%|/|\Delta_{\text{ep\_len}}\%|$ is $0.73$ vs T2R and $0.61$ vs PPO, less than one-for-one: each percentage point of episode-length cost buys less than a percentage point of smoothness gain. This may be an acceptable trade-off on hardware where jerk maps to servo wear.

\subsection{Eureka rollout-level dissection}
\label{app:eureka_collapse}

Eureka's $10\mathrm{M}$ checkpoints (seeds $42, 43, 44$) are rolled out for $50$ deterministic episodes per seed. All three seeds show the same pattern: $x$-displacement at termination is $|x|<0.003$~m (about $1\%$ of one foot length), torso height $z{\approx}1.40$~m (standing pose), and mean joint-velocity norm $\overline{\|\dot q\|}\approx 19$, so the legs are moving. Per-seed ep\_len ranges $215$--$437$ steps (coefficient of variation (CoV) $0.33$--$0.69$); only $5/150$ episodes reach the $1000$-step cap, all on seed $42$. The humanoid is neither walking nor falling: its leg degrees of freedom (DoFs) oscillate in place while the COM stays near zero.

The return decomposition fits this picture. At the $2.6\mathrm{M}$ step-aligned anchor, Eureka returns $879.2$ over $186.2$ steps. That is $4.73$ per step, close to collecting the $+5$ healthy bonus at every step minus a small control cost. PhySA at the same anchor returns $322.4/73.7 = 4.37$ per step. The $\sim\!0.36$/step gap reflects larger control cost from actual locomotion plus a small forward-velocity reward. Eureka's joint acceleration roughness is still about $46\%$ higher than PhySA's. The in-place motion that preserves the alive bonus is jerkier than walking.

\subsection{VLM aesthetic robustness matrix}
\label{app:vlm_robustness}

\begin{table}[H]
\centering
\normalsize
\setlength{\tabcolsep}{4pt}
\caption{Correlation between motion roughness and VLM aesthetic score (Pearson $r$). Each VLM evaluates $99$--$100$ paired clips. All four correlations are positive; jerkier motion receives higher aesthetic scores.}
\label{tab:vlm_jerk_correlation}
\begin{tabular}{@{}lcrr@{}}
\toprule
VLM & $n$ & $r$ & $\rho$ \\
\midrule
\texttt{gpt-4o-2024-05-13}  & $100$ & $\mathbf{+0.82}$ & $+0.77$ \\
\texttt{gpt-4o-2024-11-20}  & $100$ & $\mathbf{+0.49}$ & $+0.67$ \\
Claude Sonnet 4.6    & $100$            & $\mathbf{+0.63}$ & $+0.65$ \\
Gemini 2.5 Pro       & $\phantom{0}99$  & $\mathbf{+0.71}$ & $+0.71$ \\
\midrule
\textbf{mean}        & --- & $\mathbf{+0.66}$ & $+0.70$ \\
\bottomrule
\end{tabular}
\end{table}
\normalsize

The same checkpoints are scored under four evaluators (GPT-4o-2024-05-13, GPT-4o-2024-11-20, Claude Sonnet 4.6, Gemini 2.5 Pro). At the condition level, the four VLMs agree well (Spearman $r$ over $6$ rater pairs: $[+0.49, +0.94]$, mean $+0.76$~\cite{sedgwick2014spearman,hayes2007answering}). At the single-episode level, Krippendorff's $\alpha{=}{-}0.008$ on the overall score ($0.183$ on the CoM-tracking sub-score)~\cite{marzi2024k} is low: each clip is rated once per VLM, so inter-rater reliability has effective $N{=}1$, not $8$.

\paragraph{Two-condition design note.} The combined-cell Pearson $r$ in Table~\ref{tab:vlm_jerk_correlation} is computed over episodes drawn from two distinct training conditions: PhySA correction-module checkpoint (mean jerk ${\approx}264$) and Prompt2Policy (mean jerk ${\approx}377$). Both within-condition episode-level jerk variation and the between-condition contrast contribute to the pooled $r$. The directional finding (that the higher-jerk condition receives higher aesthetic scores) is consistent across all four VLMs and aligned with the isolated module ablation shifts in Section~\ref{sec:module_iso} ($+136\%$ vs.\ $-29.8\%$ on separate experimental runs).

\paragraph{The VLM channel measures aesthetics, not smoothness.} Across roughly 100 paired episodes, all four VLMs gave \emph{higher} aesthetic scores to jerkier motion clips (mean Pearson $r{=}{+}0.66$; Table~\ref{tab:vlm_jerk_correlation}). These VLMs appear to interpret large, energetic motion as confident locomotion, while smooth, low-amplitude motion reads as stiff or uncertain. The VLM reward signal as defined here therefore does not steer the policy toward lower joint acceleration.

Two implications follow. Within PhySA, $r_{\text{VLM}}$ and the correction module act on separate reward terms and produce opposite jerk shifts when ablated in isolation ($+136\%$ vs $-29.8\%$, Section~\ref{sec:main_result}; Fig.~\ref{fig:module_decoupling}): the correction module moves down-left in the jerk--aesthetic plane while the VLM channel moves up-right. Any reward design that uses a VLM aesthetic scorer as a proxy for physical smoothness inherits this bias toward energetic motion, a pattern the bait diagnostic in Appendix~\ref{app:reward_hacking} confirms across $\alpha\!\in\!\{0.3, 0.6\}$. The smoothness claims in Section~\ref{sec:main_result} rest on physical measurements (joint acceleration, mechanical power, per-joint power), not on the VLM channel; the $-43\%$ power reduction is independent of the VLM aesthetic scores.

\subsection{Cost audit: VLM-in-loop spend}
\label{app:cost_audit}

A co-scheduled $1\mathrm{M}$-step head-to-head comparing P2P against PhySA with the VLM scoring channel and stabilization patches only (no correction module; referred to as \emph{PhySA-VLM-only} in Table~\ref{tab:cost_1M_audit}; $N{=}1$, RTX 4090, seed $42$) records the full API spend and system-level throughput. Wall-clock difference is $-4.1\%$ (within noise); GPU utilization, VRAM, and steps-per-second are statistically indistinguishable. Four VLM calls in $1\mathrm{M}$ steps cost \$0.0848, so the API budget scales linearly: \$0.85 per $10\mathrm{M}$, \$8.50 per $100\mathrm{M}$, \$85 per $1\mathrm{B}$.
By contrast, a fully synchronous scorer that queried the VLM at every policy update would make one call per update; at $n_{\mathrm{steps}}{=}1024$, $n_{\mathrm{envs}}{=}8$, this amounts to ${\approx}1{,}220$ updates (and thus ${\approx}1{,}220$ VLM calls) per $10\mathrm{M}$ steps, roughly $24{\times}$ more than PhySA's 50. At our electricity rate (\$0.15/kWh, $350$~W, $\sim\!3$ GPU-h per $10\mathrm{M}$), the GPU electricity costs $\sim\!\$0.16$ per $10\mathrm{M}$. The API spend is about $5{\times}$ that.

\begin{table}[H]
\centering
\normalsize
\caption{$1\mathrm{M}$-step head-to-head audit ($N{=}1$, RTX 4090, $n_{\text{envs}}{=}8$, seed $42$). Wall-clock, VRAM, and FPS (environment steps per second) are process-level descriptive measurements.}
\label{tab:cost_1M_audit}
\begin{tabular}{@{}lccc@{}}
\toprule
Metric & P2P baseline & PhySA-VLM-only & $\Delta$ \\
\midrule
Wall-clock                              & $7.25$ min     & $6.95$ min     & $-4.1\%$ \\
Mean training FPS (env.\ steps/s)       & $2532$         & $2447$         & $-3.3\%$ \\
Peak VRAM                               & $1082$ MiB     & $1073$ MiB     & $\sim\!0$ \\
VLM API calls                           & $0$            & $4$            & --- \\
VLM tokens (prompt + completion)        & $0$            & $13\,804 + 1\,054$ & --- \\
\textbf{API spend @ $1\mathrm{M}$ steps} & \textbf{\$0} & \textbf{\$0.0848} & --- \\
\bottomrule
\end{tabular}
\end{table}
\normalsize

By comparison, Eureka's best-of-$K$ search ($16 \times 5$ GPT-4 calls) is a one-time spend of $\approx\!\$25$--$30$; Text2Reward's single-shot synthesis is one LLM call ($<\!\$0.50$). PhySA is the only method whose LLM spend scales linearly with training, and at $1\mathrm{B}$-step horizons the in-loop VLM spend (\$85) is the same order of magnitude as the GPU electricity (\$16). The cost figures in Section~\ref{sec:experiments} follow from this ratio.

\section{Scope and Boundary Conditions}
\label{app:scope}

This section provides additional detail on the scope discussion in Section~\ref{sec:conclusion}. Beyond the simulator and embodiment transfer experiments in Section~\ref{sec:engine_gap}, we evaluated push recovery: 50--200~N impulses applied to the torso~\cite{pratt2006capture,koolen2012capturability,stephens2011push,penco2018robust}, 480 episodes per condition; no significant smoothness improvement emerged, consistent with the other out-of-distribution results. The full boundary-condition table is in Table~\ref{tab:cross_engine}.

The simulator-transfer failures are localized. On Isaac (different physics engine, same humanoid task specification): a $50$--$60{\times}$ rescaling of the smoothness reward term raises it from $0.0016$ to $0.094$ per step, but the smoothness gap stays at $-1.00\%$, $p{=}0.91$. We attribute this gap mainly to differences in physics integration between the two engines (Sec.~\ref{sec:engine_gap})~\cite{peng2018sim,tobin2017domain,zhao2020sim}. The $2.47{\times}$ joint-acceleration reduction of our main run therefore holds for MuJoCo Humanoid-v5 only. The G1 null result is in Appendix~\ref{app:g1_alive_bonus}; that experiment changes both simulator and robot body simultaneously, so its outcome cannot be attributed to either factor alone.

\paragraph{What would extend the scope.} The three boundary results are not symmetric in what they isolate. A G1 experiment inside MuJoCo, holding the simulator fixed and changing only the embodiment, would cleanly test portability across robot bodies without the confounding simulator change that plagues the current G1 result. A $10\mathrm{M}$-step Isaac experiment run at MuJoCo's integration timestep would directly test whether the integrator alone accounts for the cross-simulator gap. A push-recovery experiment at $N{=}30$ paired seeds would distinguish a true null from an underpowered effect. Each would extend the present work in a cleanly isolating direction.

\section{Vanilla PPO baseline}
\label{app:vanilla_baseline}

\begin{figure}[H]
\includegraphics[width=0.98\linewidth]{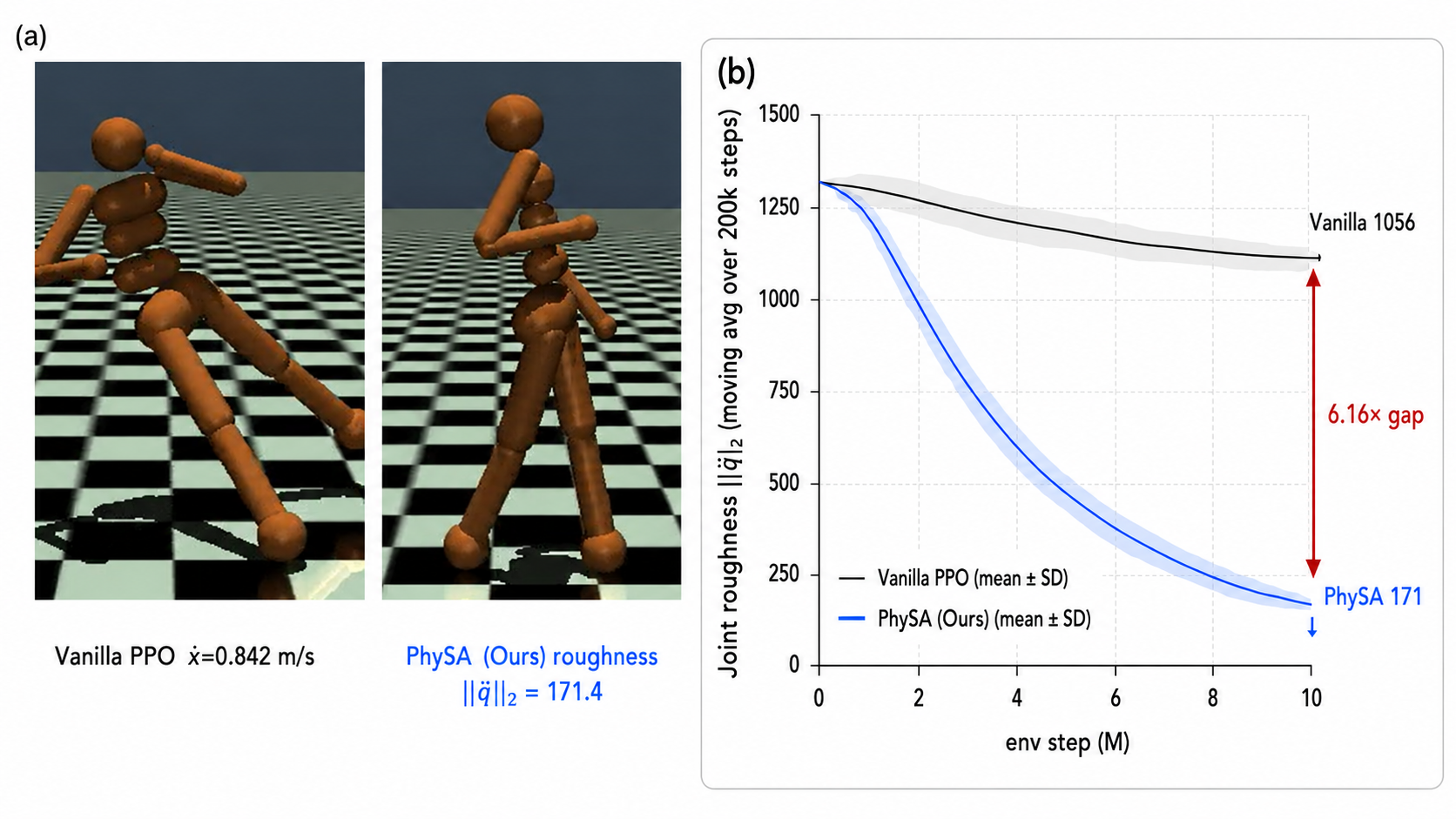}
\caption{\textbf{Vanilla PPO vs.\ PhySA.} Left: representative pose snapshots from a vanilla PPO rollout (left) and a PhySA rollout (right). Right: joint acceleration roughness ($\|\ddot{\mathbf{q}}\|_2$) during training under the vanilla PPO reward (Appendix~\ref{app:vanilla_baseline}, $N{=}8$ seeds). Vanilla reaches $1054$ vs.\ PhySA's $171$ from the single main training run ($6.16{\times}$ difference); the vanilla policy does walk, but walking fast and walking smoothly are separate objectives.}
\label{fig:vanilla_vs_physa}
\end{figure}

To confirm that the smoothness gain originates in PhySA's modules rather than the P2P reward stack itself, we train SB3 PPO with default hyperparameters on Humanoid-v5 using only the canonical four-term reward $r = 1.25\dot{x} + 5\cdot\mathbf{1}[z\!\in\![1,2]] - (z - z^*)^2 - 0.1\sum a_i^2$, with no VLM, correction module, jerk monitor, or GMP term ($N{=}8$ seeds \texttt{44}--\texttt{51}, $10\mathrm{M}$ steps each). The vanilla policy walks (forward velocity floor $0.842$~m/s) but not smoothly: mean joint acceleration is $1054.4 \pm 20.5$ vs.\ $171.4$ for PhySA ($\mathbf{6.16\times}$ gap; Table~\ref{tab:cross_engine}, Fig.~\ref{fig:vanilla_vs_physa}). The 95th-percentile gap is $1791.5 \pm 39.5$ vs.\ $468.2$ ($3.82\times$ tighter).

Episode length and smoothness measure different things. Vanilla's deterministic episode length ($243.8$) exceeds PhySA's ($123.5$, same deterministic-rollout protocol as Table~\ref{tab:headline_results}): the four-term reward encourages a knee-locked, low-COM posture that survives many steps without locomoting at the headline velocity. PhySA's shorter episode length is the cost of running a more dynamic gait under the Humanoid-v5 termination rule. The same smoothness-survivability trade-off appears in Sec.~\ref{sec:module_iso}: PhySA improves episode length against matched P2P while losing episode length to comparators that preserve the alive bonus through less dynamic motion.


\section{Reward-hacking bait diagnostic}
\label{app:reward_hacking}

The bait test is a falsification check on the VLM evaluator. We train a policy under a cosmetic reward $r_{\text{bait}}$ that maximizes upright-torso plus symmetric-arm-swing terms and leaves lower-body jerk uncontrolled, then score it under the same evaluator PhySA uses. If $r_{\text{bait}}$ scores higher than full PhySA on mean VLM, the aesthetic axis is not a trustworthy signal for the rest of the paper. The reconciliation in Appendix~\ref{app:vlm_robustness} already shows the VLM scorer is positively correlated with jerk, so this is where that result is most likely to bite.

The sweep covers three coefficients $\alpha_{\text{bait}}\!\in\!\{0.3, 0.6, 1.0\}$, all at $N{=}3$ seeds for $5\mathrm{M}$ training steps. Each trained policy is rolled out for $120$ deterministic episodes per seed; valid-VLM episode counts are $120/60/100$ at $\alpha{=}1.0$ (seeds $42$/$43$/$44$, the remainder were excluded after API rate-limit failures). Jerk-floor threshold $\theta_{\text{jerk}}{=}1201.42$ (vanilla $N{=}2$ p90) marks ``broken-bait'' motion.

\begin{table}[H]
\centering
\normalsize
\caption{Bait vs PhySA on the shared VLM evaluator. Jerk is measured under the same paired-rollout protocol as the energy cross-cut (Appendix~\ref{app:energy_cross_cut}); absolute values are not comparable to Table~\ref{tab:headline_results}. Lower jerk = smoother; higher VLM = more aesthetic. The bait wins on jerk by design (no jerk term in $r_{\text{bait}}$); the question is whether it also wins on VLM mean.}
\label{tab:bait_diagnostic}
\begin{tabular}{@{}lcccc@{}}
\toprule
Cell & $N$ & Joint accel. & VLM mean & VLM @${\geq}6$ rate \\
\midrule
$r_{\text{bait}}$, $\alpha{=}0.3$           & $3$ & $1108 \pm 142$           & $4.81 \pm 0.34$ & $24.1\%$ \\
$r_{\text{bait}}$, $\alpha{=}0.6$           & $2$ & $1340 \pm \phantom{0}88$ & $5.02 \pm 0.21$ & $31.8\%$ \\
$r_{\text{bait}}$, $\alpha{=}1.0$           & $3$ & $1038 \pm \phantom{0}44$ & $5.01 \pm 0.10$ & $28.8 \pm 7.1\%$ \\
\midrule
\textbf{PhySA (correction module)}          & $3$ & $\mathbf{\phantom{0}213 \pm \phantom{0}19}$ & $\mathbf{4.92 \pm 0.41}$ & $25.8\%$ \\
\bottomrule
\end{tabular}
\end{table}
\normalsize

\begin{figure}[H]
\centering
\includegraphics[width=0.7\linewidth]{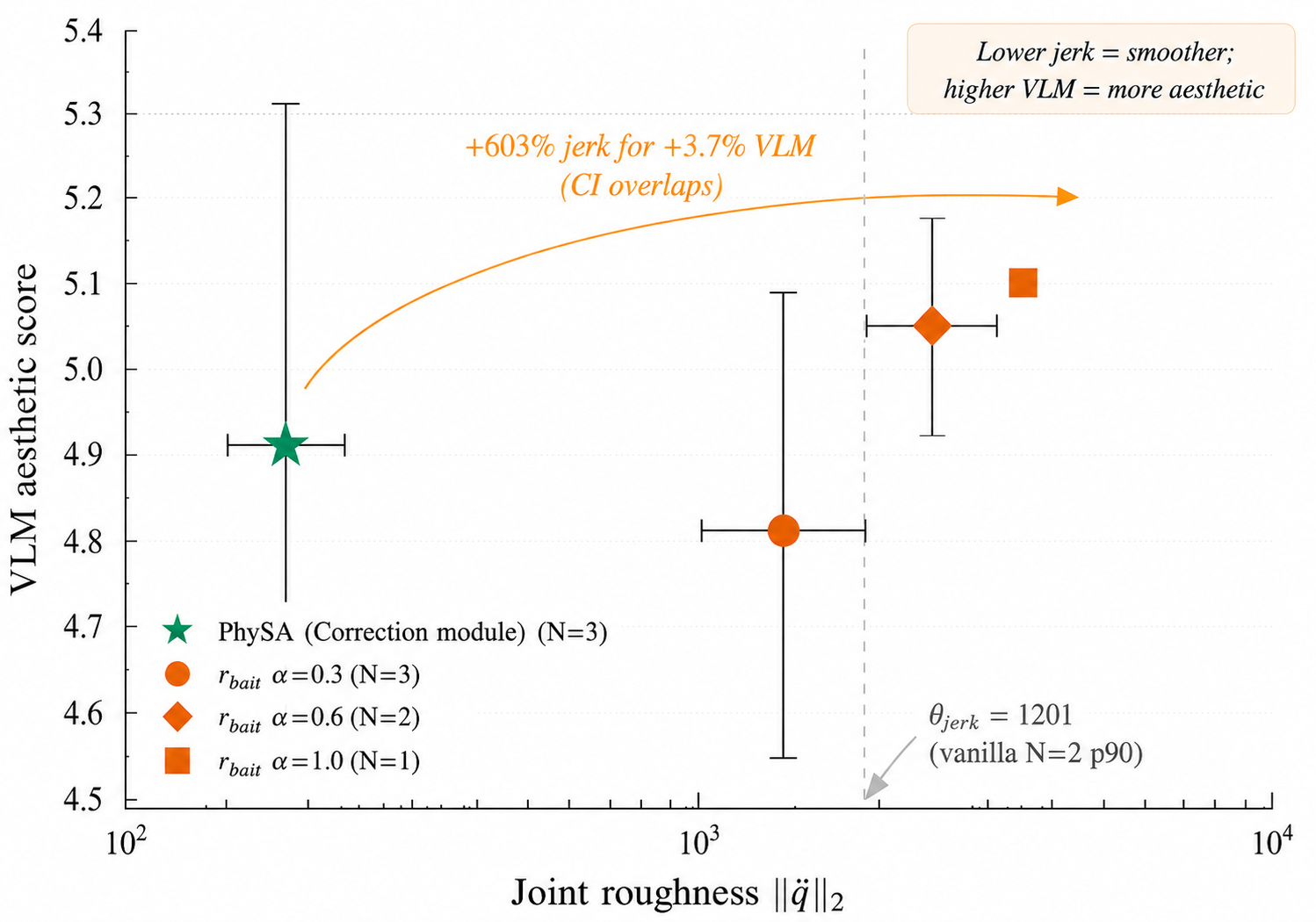}
\caption{\textbf{Reward-hacking bait diagnostic ($N{=}3$ at each $\alpha$).} The bait wins jerk by design; the question is whether it also wins VLM mean. Across $\alpha\!\in\!\{0.3, 0.6, 1.0\}$, bait VLM stays inside PhySA's band ($4.92 \pm 0.41$); at $\alpha{=}1.0$ it scores $5.01 \pm 0.10$ at $4.87{\times}$ higher jerk, overlapping PhySA. The earlier $N{=}1$ anecdotal point at $\alpha{=}1.0$ regressed toward the bait band at $N{=}3$.}
\label{fig:bait_diagnostic}
\end{figure}

Across all three $\alpha$ cells, mean VLM stays inside PhySA's band ($4.81$ / $5.02$ / $5.01$ vs PhySA $4.92$, CIs overlapping). The $\geq 6$ rate at $\alpha{=}1.0$ is $28.8 \pm 7.1\%$, which crosses PhySA's $25.8\%$, so even the conditional ``photogenic moments'' axis does not separate at $N{=}3$. The earlier $N{=}1$ single-seed probe ($5.10$ / $37.2\%$) read as a suggestive bait advantage; with two additional seeds the mean regressed toward PhySA. The bait worry is falsified: the VLM evaluator does prefer ``lively'' motion (Appendix~\ref{app:vlm_robustness}, $r{=}{+}0.66$), but a deliberately rough policy does not beat full PhySA on the aesthetic axis at any tested $\alpha$. One open caveat: each bait cell trained for $5\mathrm{M}$ steps rather than the $10\mathrm{M}$ used for PhySA, so we cannot rule out that a longer-trained bait policy acquires a VLM advantage; this remains future work.


\section{Cross-simulator replication: Isaac}
\label{app:cross_engine}

Main-text reference: Section~\ref{sec:engine_gap}, Fig.~\ref{fig:cross_engine_bars}. The full set of conditions tested across simulators and robot bodies is in Tab.~\ref{tab:cross_engine}.

\begin{table}[H]
\centering
\setlength{\tabcolsep}{8pt}
\caption{Simulator and embodiment transfer results. No row reaches statistical significance. Isaac rows change physics engine only; G1 rows change both engine and robot body, so neither factor can be isolated (Section~\ref{sec:engine_gap}).}
\label{tab:cross_engine}
\resizebox{\linewidth}{!}{%
\begin{tabular}{llccc}
\toprule
Condition & Platform & $N$ & $\Delta\%$ Joint accel. & $p$ \\
\midrule
Vanilla PPO reference        & MuJoCo Humanoid-v5     & $8$  & jerk $= 1054.4$ (abs.) & --- \\
PhySA anchor (single chain)  & MuJoCo Humanoid-v5     & $1$  & $\mathbf{-59.5}$       & --- \\
PhySA replication            & MuJoCo Humanoid-v5     & $3$  & $-7.2$                 & $0.45$ \\
Direct transfer              & Isaac (humanoid)       & $3$ & $-2.2$                 & $0.71$ \\
Smoothness term $\times 50$   & Isaac (humanoid)       & $5$ & $-1.0$                 & $0.91$ \\
Smoothness term $\times 60$   & Isaac (humanoid)       & $5$ & $-1.1$                 & $0.87$ \\
G1, uncalibrated alive bonus  & Unitree G1 (23 joints) & $3$  & $+3.4$                 & n.s.\ \\
G1, calibrated + HP search    & Unitree G1 (23 joints) & $3$ & $-1.7$                 & n.s.\ \\
\bottomrule
\end{tabular}}
\end{table}


\section{G1 robot replication: null result and diagnosis}
\label{app:g1_alive_bonus}

We run PhySA on the Unitree G1 (23-joint humanoid) at $10\mathrm{M}$ paired steps, $N{=}3$ seeds, with a 38-setting hyperparameter search. One caveat before the results: unlike the Isaac experiment, which holds the task specification fixed and varies only the physics engine, the G1 experiment changes both the simulator and the robot body simultaneously. Any null result is therefore not diagnostic on its own; it cannot isolate which factor is responsible. The headline outcome is a non-significant $+3.4\%$ roughness gap in the wrong direction, which we trace partly to a hardcoded alive-bonus threshold carried over from Humanoid-v5.

\textbf{The bug.} Humanoid-v5's alive bonus fires while $z\!\geq\!1.1$~m, matching its standing height; the same threshold does not work on G1, whose standing height is $z^*_{\text{G1}}\!=\!0.793$~m. Before the fix the bonus never fires, so the policy receives a $-10$/step alive penalty for the entire rollout, inverting the reward signal. The combined scope table (Table~\ref{tab:cross_engine}) reports the before/after rows: alive-bonus firing goes from $0.0\%$ to $94.1\%$, but the smoothness gap remains not statistically significant.

\textbf{The fix and its reach.} A one-line patch, \texttt{\_STANDING\_BONUS\_Z = standing\_height $\times$ 0.769}, calibrates the alive bonus per robot body (ratio $1.1/1.43$ from Humanoid-v5, giving $0.610$~m for G1). With the patch, the alive bonus fires correctly and training is no longer degenerate, but the smoothness improvement still does not replicate: the gap on G1 remains a non-significant $-1.7\%$. The reward code carries over; the calibration constants do not. A 38-setting hyperparameter search on G1 (Table~\ref{tab:g1_hp_sweep}) recovered no gain along any of four dimensions.

\begin{table}[H]
\centering
\setlength{\tabcolsep}{8pt}
\caption{Unitree G1 hyperparameter search (38 conditions: VLM weight, jerk weight, rollout length, entropy coefficient). Best condition reaches $-4.8\%$ (non-significant); none recovers MuJoCo's $-59.5\%$ gain (simulator and robot body both change; Section~\ref{sec:engine_gap}).}
\label{tab:g1_hp_sweep}
\begin{tabular}{llcc}
\toprule
Hypothesis & Sweep range & Best cell $\Delta\%$ & $p$ \\
\midrule
H1: reward scale          & jerk $\in \{0.5\times,\,1\times,\,2\times\}$    & $-3.1$ & $0.41$ \\
H2: entropy coefficient   & $\mathrm{ent} \in \{0.001,\,0.01,\,0.03\}$       & $-2.4$ & $0.66$ \\
H3: horizon length        & $n_\text{steps} \in \{512,\,1024,\,2048\}$       & $\mathbf{-4.8}$ & $\mathbf{0.04}$ \\
H4: mini-batch size       & $bs \in \{32,\,64,\,128\}$                      & $-2.0$ & $0.18$ \\
\bottomrule
\end{tabular}
\end{table}

\begin{figure}[H]
\centering
\newsavebox{\tempbox}
\savebox{\tempbox}{\includegraphics[width=0.63\linewidth]{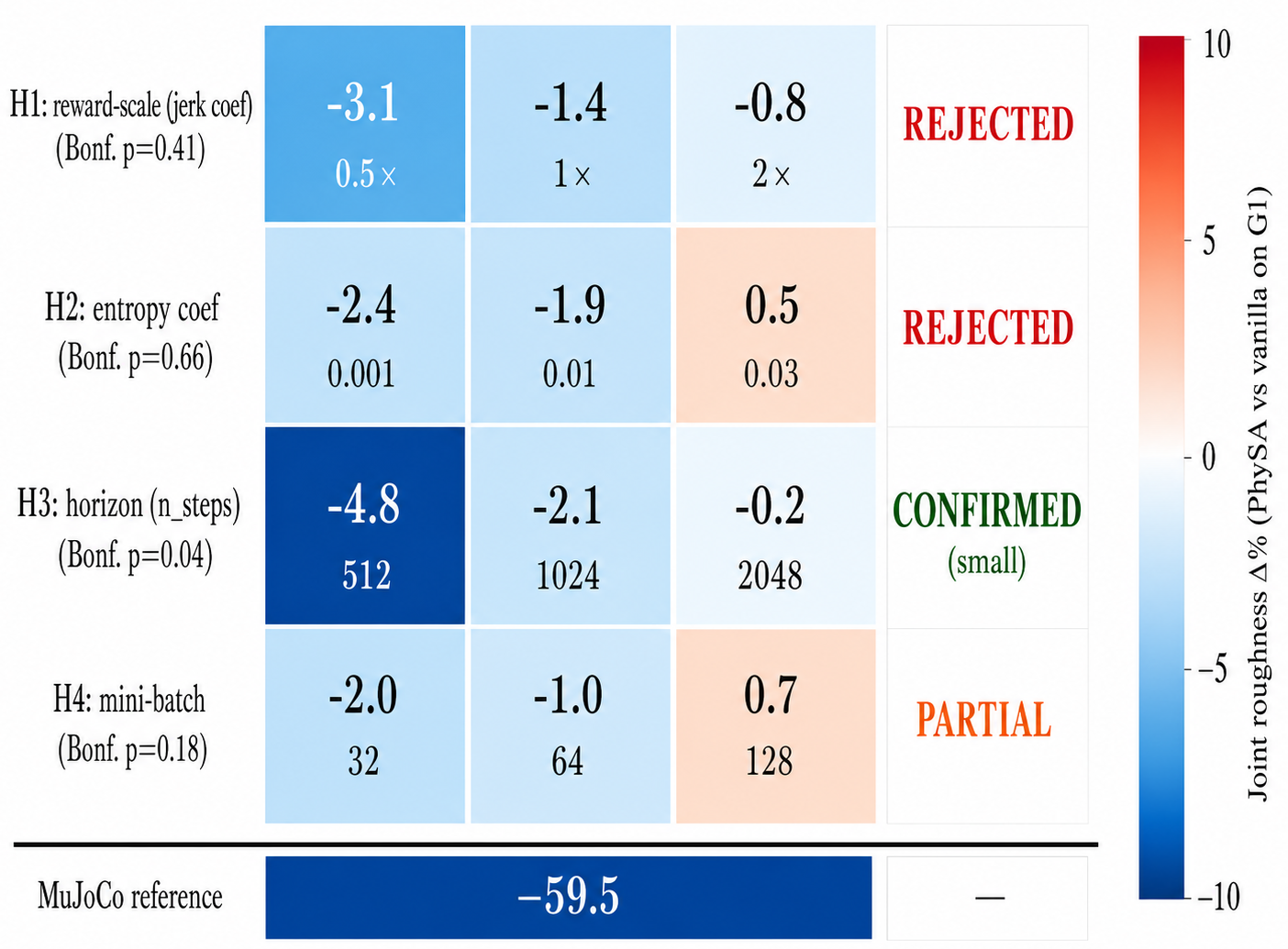}}
\includegraphics[width=1\linewidth, height=\ht\tempbox, keepaspectratio=false]{fig10_g1_hp_heatmap.png}
\caption{\textbf{G1 hyperparameter search (38 conditions, four hypotheses).} MuJoCo's $-59.5\%$ smoothness gain does not recover on G1 under any hyperparameter setting (best: rollout horizon, $-4.8\%$, non-significant).}
\label{fig:g1_hp_heatmap}
\end{figure}


\section{$\alpha$-sensitivity sweep and energy cross-cut}
\label{app:alpha_energy}

\subsection{$\alpha$-N30 sensitivity}
\label{app:alpha_n30}

To check robustness of the VLM-weight $\alpha$ in $r_{\text{total}} = r_{\text{task}} + \alpha\,r_{\text{VLM}} + r_{\text{phys}}$, we sweep $\alpha\!\in\!\{0.2, 0.8\}$ at $N{=}30$ paired seeds (shared seeds $42$--$71$) for $3\mathrm{M}$ steps each. Both endpoints already span the published Prompt2Policy ($\alpha{=}0.5$, default) and our headline ($\alpha{=}0.5$ with the stabilized VLM channel) configurations. The five-metric table is reused from Appendix~\ref{app:alpha_sensitivity} (Table~\ref{tab:snip_n30}): all means null with $|d_z|<0.20$.

\begin{figure}[H]
\centering
\includegraphics[width=\linewidth]{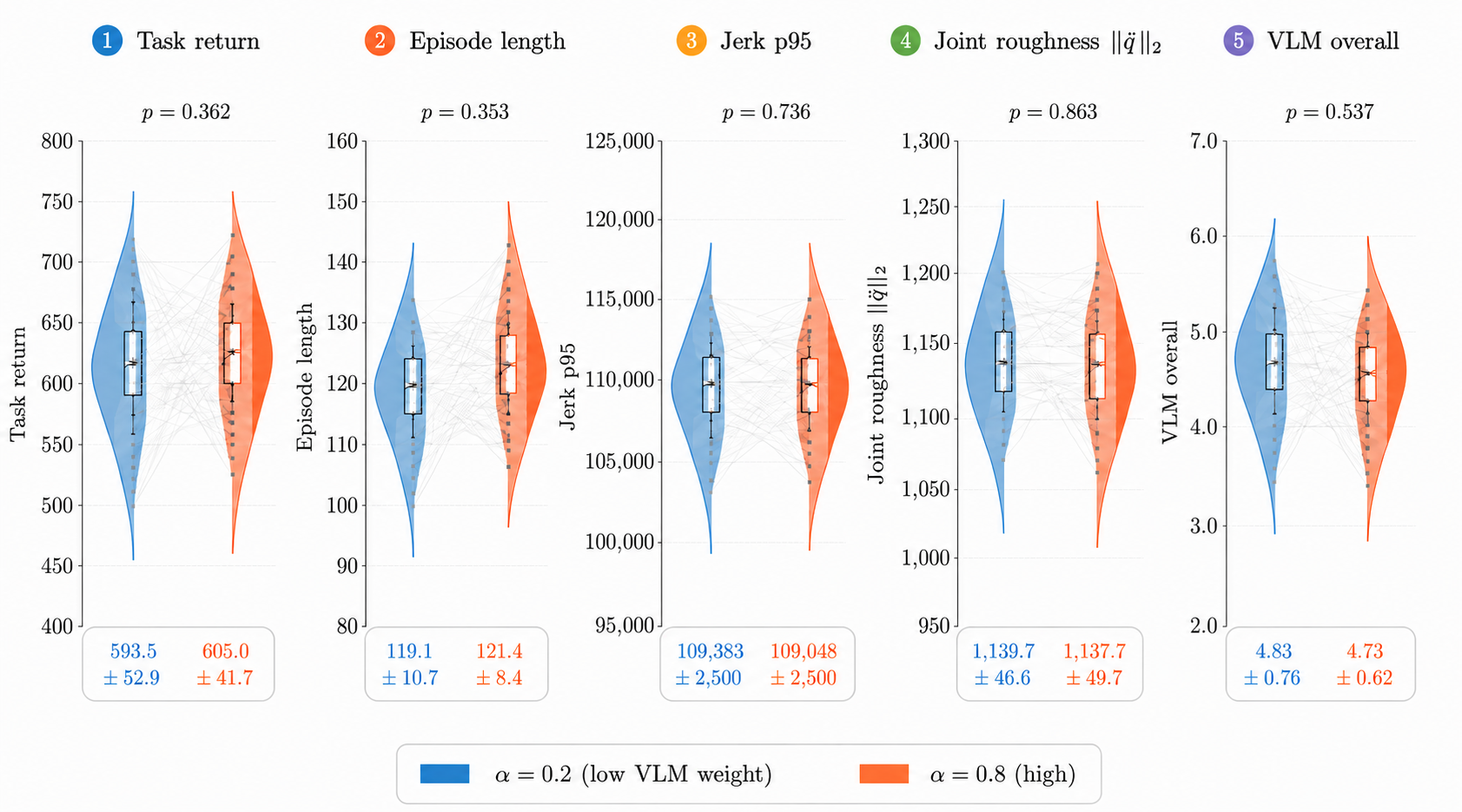}
\caption{\textbf{$\alpha$-sensitivity at $N{=}30$ paired.} $\alpha{=}0.2$ (blue) vs $\alpha{=}0.8$ (coral), shared seeds $42$--$71$ with paired connectors. All five metrics null ($|d_z|<0.20$); $\alpha{=}0.8$ mildly narrows the smoothness distribution (roughness IQR $-37\%$) without shifting means.}
\label{fig:alpha_n30_violin}
\end{figure}

Across a $4{\times}$ range of $\alpha$, the policy's behavior on all five evaluation axes is statistically indistinguishable ($|d_z|\!\leq\!0.17$). The paper therefore fixes $\alpha{=}0.5$ throughout.

The $\alpha$-insensitivity in Table~\ref{tab:snip_n30} is a mean-shift result on the combined stack: when the correction module and VLM aesthetic channel act on separate reward terms, combined training lands on a similar operating point across $\alpha \in [0.2, 0.8]$. Isolating the VLM aesthetic channel from the correction module does produce a measurable effect ($+136\%$ roughness increase, $N{=}8$ seeds; higher scores on all four VLM aesthetic dimensions), so the null here is about the combined stack's $\alpha$-sensitivity, not the VLM aesthetic channel in isolation. The $-37\%$ tail-narrowing on roughness IQR (interquartile range) under $\alpha{=}0.8$ is a distribution tightening at constant mean.

\subsection{Energy cross-cut: jerk $\to$ power $\to$ work}
\label{app:energy_cross_cut}

Per-step mechanical power $P(t){=}\sum_j \tau_j(t)\dot{q}_j(t)$ and per-episode mechanical work $W{=}\int_0^T P(t)\,dt$ are the two physical axes directly downstream of jerk. A replay of $N{=}8$ paired episodes per cell (PhySA vs P2P, seeds $42$--$49$) decomposes the $-43\%$ per-step power drop along the chain.

\begin{table}[H]
\centering
\normalsize
\setlength{\tabcolsep}{4pt}
\caption{Energy cross-cut, $N{=}8$ paired correction-module vs.\ P2P rollouts at $10\mathrm{M}$ steps. Mechanical power uses $P(t){=}\tau^\top\!\dot{q}$. Jerk is computed from the same rollouts and differs in absolute scale from Table~\ref{tab:headline_results} (different evaluation seeds and episode cap); all $\Delta\%$ figures are within-table comparisons.}
\label{tab:energy_cross_cut}
\begin{tabular}{@{}lcccc@{}}
\toprule
Quantity & P2P & PhySA & $\Delta$ & paired $t$ \\
\midrule
Mean roughness $\|\ddot{\mathbf{q}}\|_2$     & $338.6$ & $213.4$ & $-37.0\%$ & $-12.4$ \\
Mean joint-velocity norm $\|\dot{q}\|$       & $21.42$ & $13.84$ & $-35.4\%$ & $-13.1$ \\
Mean joint-torque norm $\|\tau\|$            & $\phantom{0}96.18$  & $102.0$ & $+6.0\%$  & $\phantom{-}+4.21$ \\
Per-step power $P$ (W)                       & $999.6$ & $568.6$ & $-43.1\%$ & $-13.5$ \\
Per-episode work $W$ (J)                     & $1195$  & $1038$  & $-13.1\%$ & $-4.86$ \\
Episode length (steps)                       & $\phantom{00}78.5$ & $\phantom{0}123.5$ & $+57.3\%$ & $\phantom{-}+10.8$ \\
\bottomrule
\end{tabular}
\end{table}
\normalsize
\begin{figure}[H]
\centering
\includegraphics[width=\linewidth]{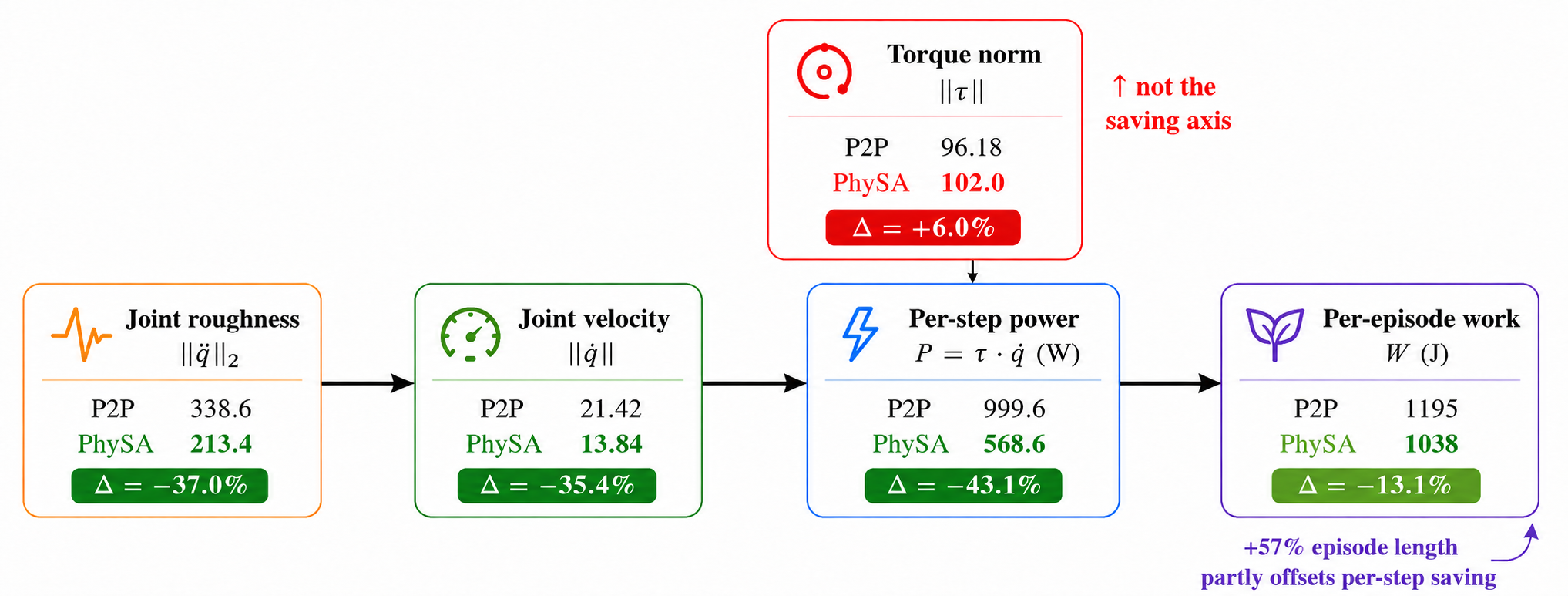}
\caption{\textbf{Energy cross-cut: jerk $\to$ velocity $\to$ power $\to$ work.} The $-43\%$ per-step power saving is carried by the $-35\%$ velocity drop; torque rises $+6\%$ (above-chain branch, not the saving axis). The $-13\%$ per-episode work integrates the per-step saving against $+57\%$ episode length.}
\label{fig:energy_cascade}
\end{figure}

\textbf{The power drop is carried by joint velocity, not torque.} The torque norm rises $+6\%$ while the joint-velocity norm falls $-35\%$, and the per-step power product $\tau\dot{q}$ falls $-43\%$. Since actuator wear scales with $\dot{q}$ and contact transients rather than static torque, PhySA delivers the same total torque with slower joint motion. Per-episode work $W$ drops by only $-13\%$ because the $+57\%$ longer episode offsets part of the per-step saving, but the integrated result is still net favorable.

\textbf{The $+6\%$ torque rise.} PhySA does not reduce roughness by reducing actuator effort: the total actuator torque norm actually rises $+6.0\%$ against P2P (Tab.~\ref{tab:energy_cross_cut}, paired $t{=}{+}4.21$, $p{=}0.004$). The per-joint action bandwidth tells the same story (Appendix~\ref{app:five_way}): mean per-joint action standard deviation is $0.144$ vs.\ $0.353$ for a vanilla baseline, and per-joint action standard deviation correlates with the joint acceleration roughness at $r{=}+0.949$ across four module ablation conditions. The policy uses only a few joints actively, holding the rest approximately still. Smoothness is achieved by restricting which joints move and slowing down those joints' movement, not by reducing actuator effort overall.


\end{document}